\newcommand{\s}[2]{#1}  % For arXiv version
\documentclass{article}
\usepackage{nips15submit_e,times}
\nipsfinalcopy

\usepackage{times}
\usepackage{graphicx} % more modern
\usepackage{subfigure} 

\s{\usepackage{natbib}}{\usepackage[numbers]{natbib}}
\usepackage{algorithm}
\usepackage{algorithmic}
\usepackage{multicol}

\usepackage[hidelinks]{hyperref}

\usepackage{amsbsy}

\newcommand{\vect}[1]{\mathbf{#1}}

\newcommand{\vh}[0]{\vect{h}}
\newcommand{\vv}[0]{\vect{v}}

\newcommand{\vx}[0]{\vect{x}}
\newcommand{\vy}[0]{\vect{y}}
\newcommand{\vz}[0]{\vect{z}}

\newcommand{\vmu}[0]{\boldsymbol{\mu}}
\newcommand{\vupsilon}[0]{\boldsymbol{\upsilon}}

\renewcommand{\l}{^{(l)}}
\newcommand{\x}{\mathbf{x}}

\newcommand{\y}{\mathbf{y}}

\newcommand{\f}[1]{f^{(#1)}}
\newcommand{\g}[1]{g^{(#1)}}

\renewcommand{\u}[1]{\mathbf{u}^{(#1)}}
\newcommand{\h}[1]{\vh^{(#1)}}

\newcommand{\tildeh}[1]{\tilde{\vh}^{(#1)}}
\newcommand{\z}[1]{\vz^{(#1)}}
\newcommand{\hz}[1]{\hat{\vz}^{(#1)}}
\newcommand{\tz}[1]{\tilde{\vz}^{(#1)}}

\newcommand{\W}[1]{\mathbf{W}^{(#1)}}
\newcommand{\V}[1]{\mathbf{V}^{(#1)}}

\begin{document} 

%\twocolumn[
%\title{Lateral Connections in Denoising Autoencoders Support
%Supervised and Semi-Supervised Learning}

\title{Semi-Supervised Learning with Ladder Networks}

% unless you've provided the [accepted] option to the icml2015
\s{\author{
Antti Rasmus \\ The Curious AI Company, Finland \And
Harri Valpola \\ The Curious AI Company, Finland \And
Mikko Honkala \\ Nokia Labs, Finland \And
Mathias Berglund \\ Aalto University \& The Curious AI Company, Finland \And
Tapani Raiko \\ Aalto University \& The Curious AI Company, Finland}}
{\author{%
Antti Rasmus and Harri Valpola \\ The Curious AI Company, Finland \And
Mikko Honkala \\ Nokia Labs, Finland \And
Mathias Berglund and Tapani Raiko \\ Aalto University, Finland \& The Curious AI Company, Finland}}

%\icmlkeywords{semi-supervised learning}

%\vskip 0.3in
%]
\maketitle

\begin{abstract} 
We combine supervised learning with unsupervised learning in deep neural networks.
The proposed model is trained to simultaneously minimize the sum of supervised 
and unsupervised cost functions by 
backpropagation, avoiding the need for layer-wise pre-training.
Our work builds on 
the Ladder network proposed by \citet{valpola2015ladder}, which we extend by combining
the model with supervision.
We show that the resulting model reaches state-of-the-art performance in semi-supervised
MNIST and CIFAR-10 classification, in addition to permutation-invariant MNIST classification
with all labels.
\end{abstract}

% % % % % % % % % % % % % % 
\section{Introduction}
% % % % % % % % % % % % % % 

\s{
In this paper, we introduce an unsupervised learning method that fits well
with supervised learning. The idea of using unsupervised learning to
complement supervision is not new.
Combining an auxiliary task to help train a neural network was proposed by
\citet{suddarth1990rule}. By sharing the hidden representations 
among more than one task, the network generalizes better.
There are multiple choices for the unsupervised task, for example,
reconstruction of the inputs at every level of the model
\citep[e.g.,][]{Ranzato2008semi}
or classification of each input sample into its
own class \citep{dosovitskiy2014discriminative}.

Although some methods have been able to simultaneously apply both
supervised and unsupervised learning
\citep{Ranzato2008semi,goodfellow2013multi},
often these unsupervised auxiliary tasks are only applied as pre-training,
followed by normal supervised learning \citep[e.g.,][]{hinton2006reducing}.
In complex tasks there is often much more structure in the inputs
than can be represented, and unsupervised learning cannot, by definition,
know what will be useful for the task at hand. Consider, for instance, the
autoencoder approach applied to natural images: an auxiliary decoder
network tries to reconstruct the original input from the internal
representation. The autoencoder will try to preserve all the details needed for
reconstructing the image at pixel level, even though classification is
typically invariant to all kinds of transformations which do not preserve
pixel values. Most of the information required for pixel-level reconstruction
is irrelevant and takes space from the more relevant invariant features
which, almost by definition, cannot alone be used for reconstruction.

Our approach follows \citet{valpola2015ladder}, who proposed a Ladder network
where the auxiliary task is to denoise representations at every level of
the model. The model structure is an autoencoder with skip connections from
the encoder to decoder and the learning task is similar to that in
denoising autoencoders but applied to every layer, not just the inputs.
The skip connections relieve the pressure to represent details in the
higher layers of the model because, through the skip connections,
the decoder can recover any details discarded by the encoder.
Previously, the Ladder network has only been demonstrated in unsupervised
learning \citep{valpola2015ladder,Rasmus15arxiv} but we now combine
it with supervised learning.

The key aspects of the approach are as follows:

\textbf{Compatibility with supervised methods}. The unsupervised part focuses on relevant details
found by supervised learning. Furthermore, it can be added to existing feedforward neural networks, for example
multi-layer perceptrons (MLPs) or convolutional neural networks (CNNs) (Section~\ref{sec:implementation}).
We show that we can take a state-of-the-art supervised
learning method as a starting point and improve the network further by
adding simultaneous unsupervised learning (Section~\ref{sec:experiments}).

\textbf{Scalability resulting from local learning}. In addition to a supervised
learning target on the top layer, the model has local unsupervised learning
targets on every layer, making it suitable
for very deep neural networks. We demonstrate this with two deep supervised network architectures.

\textbf{Computational efficiency}. The encoder part of the model corresponds to normal supervised
learning. Adding a decoder, as proposed in this paper, approximately triples the
computation during training but
not necessarily the training time since the same result can be achieved faster
through the better utilization
of the available information. Overall, computation per update scales similarly to
whichever supervised learning approach is used, with a small multiplicative
factor.

As explained in Section~\ref{sec:derivation}, the skip connections and
layer-wise unsupervised targets effectively turn autoencoders into
hierarchical latent variable models which are known to be well suited
for semi-supervised learning. Indeed, we obtain state-of-the-art results in
semi-supervised learning in the MNIST, permutation invariant MNIST and CIFAR-10
classification tasks (Section~\ref{sec:experiments}). However, the
improvements are not limited to semi-supervised settings: for the
permutation invariant MNIST task, we also achieve a new record with
the normal full-labeled setting.%
\footnote{Preliminary results on
the full-labeled setting on a permutation invariant MNIST task were
reported in a short early version of this paper \citep{rasmus2015icml}.
Compared to that, we have added noise to all layers of the model and
further simplified the denoising function $g$. This further improved
the results.}
% END OF ARXIV VERSION
}{
% START OF NIPS VERSION
In this paper, we introduce an unsupervised learning method that fits well
with supervised learning.
% The idea of using unsupervised learning to
% complement supervision is not new.
Combining an auxiliary task to help train a neural network was proposed by
\citet{suddarth1990rule}.
% By sharing the hidden representations 
% among more than one task, the network generalizes better.
There are multiple choices for the unsupervised task, for example,
reconstruction of the inputs at every level of the model
\citep[e.g.,][]{Ranzato2008semi}
or classification of each input sample into its
own class \citep{dosovitskiy2014discriminative}.

Although some methods have been able to simultaneously apply both
supervised and unsupervised learning
\citep{Ranzato2008semi,goodfellow2013multi},
often these unsupervised auxiliary tasks are only applied as pre-training,
followed by normal supervised learning \citep[e.g.,][]{hinton2006reducing}.
In complex tasks there is often much more structure in the inputs
than can be represented, and unsupervised learning cannot, by definition,
know what will be useful for the task at hand. Consider, for instance, the
autoencoder approach applied to natural images: an auxiliary decoder
network tries to reconstruct the original input from the internal
representation. The autoencoder will try to preserve all the details needed for
reconstructing the image at pixel level, even though classification is
typically invariant to all kinds of transformations which do not preserve
pixel values.
% Most of the information required for pixel-level reconstruction
% is irrelevant and takes space from the more relevant invariant features
% which, almost by definition, cannot alone be used for reconstruction.

Our approach follows \citet{valpola2015ladder}, who proposed a Ladder network
where the auxiliary task is to denoise representations at every level of
the model. The model structure is an autoencoder with skip connections from
the encoder to decoder and the learning task is similar to that in
denoising autoencoders but applied to every layer, not just the inputs.
The skip connections relieve the pressure to represent details in the
higher layers of the model because, through the skip connections,
the decoder can recover any details discarded by the encoder.
Previously, the Ladder network has only been demonstrated in unsupervised
learning \citep{valpola2015ladder,Rasmus15arxiv} but we now combine
it with supervised learning.

The key aspects of the approach are as follows:

\textbf{Compatibility with supervised methods}. The unsupervised part focuses on relevant details
found by supervised learning. Furthermore, it can be added to existing feedforward neural networks, for example
multi-layer perceptrons (MLPs) or convolutional neural networks (CNNs).
%(Section~\ref{sec:implementation}).
% We show that we can take a state-of-the-art supervised
% learning method as a starting point and improve the network further by
% adding simultaneous unsupervised learning (Section~\ref{sec:experiments}).

\textbf{Scalability resulting from local learning}. In addition to a supervised
learning target on the top layer, the model has local unsupervised learning
targets on every layer, making it suitable
for very deep neural networks. We demonstrate this with two deep supervised network architectures.

\textbf{Computational efficiency}. The encoder part of the model corresponds to normal supervised
learning. Adding a decoder, as proposed in this paper, approximately triples the
computation during training but
not necessarily the training time since the same result can be achieved faster
through the better utilization
of the available information. Overall, computation per update scales similarly to
whichever supervised learning approach is used, with a small multiplicative
factor.

As explained in Section~\ref{sec:derivation}, the skip connections and
layer-wise unsupervised targets effectively turn autoencoders into
hierarchical latent variable models which are known to be well suited
for semi-supervised learning. Indeed, we obtain state-of-the-art results in
semi-supervised learning in the MNIST, permutation invariant MNIST and CIFAR-10
classification tasks (Section~\ref{sec:experiments}). However, the
improvements are not limited to semi-supervised settings: for the
permutation invariant MNIST task, we also achieve a new record with
the normal full-labeled setting.%
% \footnote{Preliminary results on
% the full-labeled setting on permutation invariant MNIST task were
% reported in a short early version of this paper \citep{rasmus2015icml}.
% Compared to that, we have added noise to all layers of the model and
% further simplified the denoising function $g$. This further improved
% the results.}
For a longer version of this paper with more complete descriptions, please see \cite{rasmus2015semi}.
}

% % % % % % % % % % % % % % 
\section{Derivation and justification}
% % % % % % % % % % % % % % 
\label{sec:derivation}

Latent variable models are an attractive approach to semi-supervised
learning because they can combine supervised and unsupervised learning
in a principled way. The only difference is whether the class
labels are observed or not. This approach was taken, for instance,
by \citet{goodfellow2013multi} with their multi-prediction deep Boltzmann
machine. A particularly attractive property of hierarchical latent variable
models is that they can, in general, leave the details for the lower levels
to represent, allowing higher levels to focus on more invariant, abstract
features that turn out to be relevant for the task at hand.

The training process of latent variable models can typically be split into
inference and learning,
that is, finding the posterior probability of the unobserved latent variables
and then updating the underlying probability model to fit the
observations better. For instance, in the expectation-maximization (EM) algorithm,
the E-step corresponds to finding the expectation of the latent variables
over the posterior distribution assuming the model fixed and the M-step then
maximizes the underlying probability model assuming the expectation fixed.

The main problem with latent variable models is how
to make inference and learning efficient. Suppose there are layers $l$
of latent variables $\z l$.
Latent variable models often represent the probability distribution
of all the variables explicitly as a product of terms, such as
$p(\z l \mid \z {l+1})$ in directed graphical models. The inference process and
model updates are then derived from Bayes' rule, typically as some kind
of approximation. The inference is often iterative as it is generally
impossible to solve the resulting equations in a closed form as a function
of the observed variables.

There is a close connection between denoising and probabilistic modeling.
On the one hand,
given a probabilistic model, you can compute the optimal denoising. Say you want
to reconstruct a latent $z$ using a prior $p(z)$ and an observation
$\tilde{z}=z+\textrm{noise}$. We first compute the posterior distribution
$p(z \mid \tilde{z})$, and use its center of gravity as the reconstruction $\hat{z}$.
One can show that this minimizes the expected denoising cost $(\hat{z} - z)^2$.
On the other hand, given a denoising function, one can draw samples from the
corresponding distribution by creating a Markov chain that alternates between
corruption and denoising \citep{bengio2013gsn}.

\citet{valpola2015ladder} proposed the Ladder network, where the inference process
itself can be learned by using the principle of denoising, which has been used
in supervised learning \citep{sietsma1991creating}, denoising autoencoders
(dAE) \citep{vincent2010stacked}, and denoising source
separation (DSS) \citep{sarela2005denoising} for complementary tasks.
In dAE, an autoencoder is trained to reconstruct the original observation $\x$
from a corrupted version $\tilde \x$. Learning is based simply on minimizing
the norm of the difference of the original $\x$ and its reconstruction
$\hat \x$ from the corrupted $\tilde \x$; that is the cost is $\Vert \hat{\x} - \x \Vert^2$.

While dAEs are normally only trained to denoise
the observations, the DSS framework is based on the idea of using denoising
functions $\hat{\vz} = g(\vz)$ of the latent variables $\vz$ to
train a mapping $\vz = f(\x)$ which models the likelihood
of the latent variables as a function of the observations. The cost function is
identical to that used in a dAE except that the latent variables $\vz$ replace the
observations $\vx$; that is, the cost is $\Vert \hat{\vz} - \vz \Vert^2$.
The only thing to keep in mind is that $\vz$ needs to be normalized somehow as
otherwise the model has a trivial solution at $\vz = \hat{\vz} =$ constant.
In a dAE, this cannot happen as the model cannot change the input $\x$.

Figure~\ref{fig:denois} \s{}{(left) }depicts the optimal denoising function $\hat{z} = g(\tilde z)$
for a one-dimensional bimodal distribution, which could be the distribution of a latent
variable inside a larger model. The shape of the denoising function depends on the
distribution of $z$ and the properties
of the corruption noise. With no noise at all, the optimal denoising function would be
the identity function. In general, the denoising function pushes the values towards higher
probabilities, as shown by the green arrows.

Figure~\ref{fig:ladder} \s{}{(right) }shows the structure of the Ladder
network. Every layer contributes
to the cost function a term $C_d^{(l)}=\Vert \z l - \hz l \Vert^2$ which trains the layers above
(both encoder and decoder) to learn the denoising function $\hz l = g \l(\tz l, \hz {l+1})$
which maps the corrupted $\tz l$ onto the denoised estimate $\hz l$.
As the estimate $\hz l$ incorporates all prior knowledge about $\vz$, the same cost
function term also trains the encoder layers below to find cleaner features which
better match the prior expectation.

Since the cost function needs both the clean $\z l$ and corrupted $\tz l$, during training the
encoder is run twice: a clean pass for $\z l$ and a corrupted pass for $\tz l$.
Another feature which differentiates the Ladder network from regular dAEs is that
each layer has a skip connection between the encoder and decoder. This feature mimics
the inference structure of latent variable models and makes it possible for the
higher levels of the network to leave some of the details for lower levels to represent.
\citet{Rasmus15arxiv} showed that such skip connections allow dAEs to focus on abstract
invariant features on the higher levels, making the Ladder network a good fit with
supervised learning that can select which information is relevant for the task at hand.

\s{
\begin{figure}[tbp]
\begin{center}
  \includegraphics[width=0.7\columnwidth]{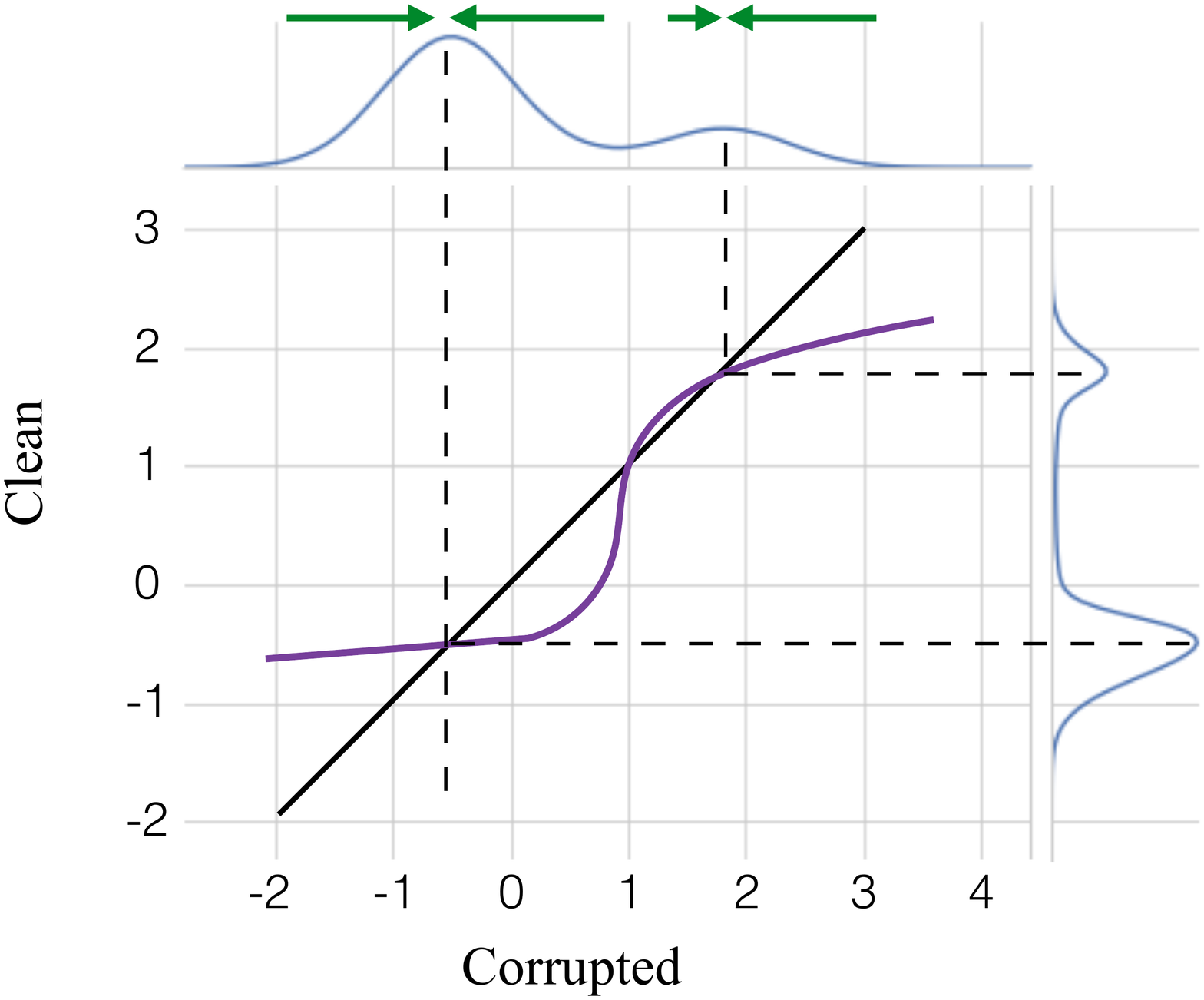}
\caption{A depiction of an optimal denoising function for a bimodal distribution. The
input for the function is the corrupted value (x axis) and the target is the clean
value (y axis). The denoising function moves values towards higher probabilities as
show by the green arrows.}
\label{fig:denois}
\end{center}
\end{figure}

\begin{figure}[tbp]
\begin{center}
  \includegraphics[width=0.9\columnwidth]{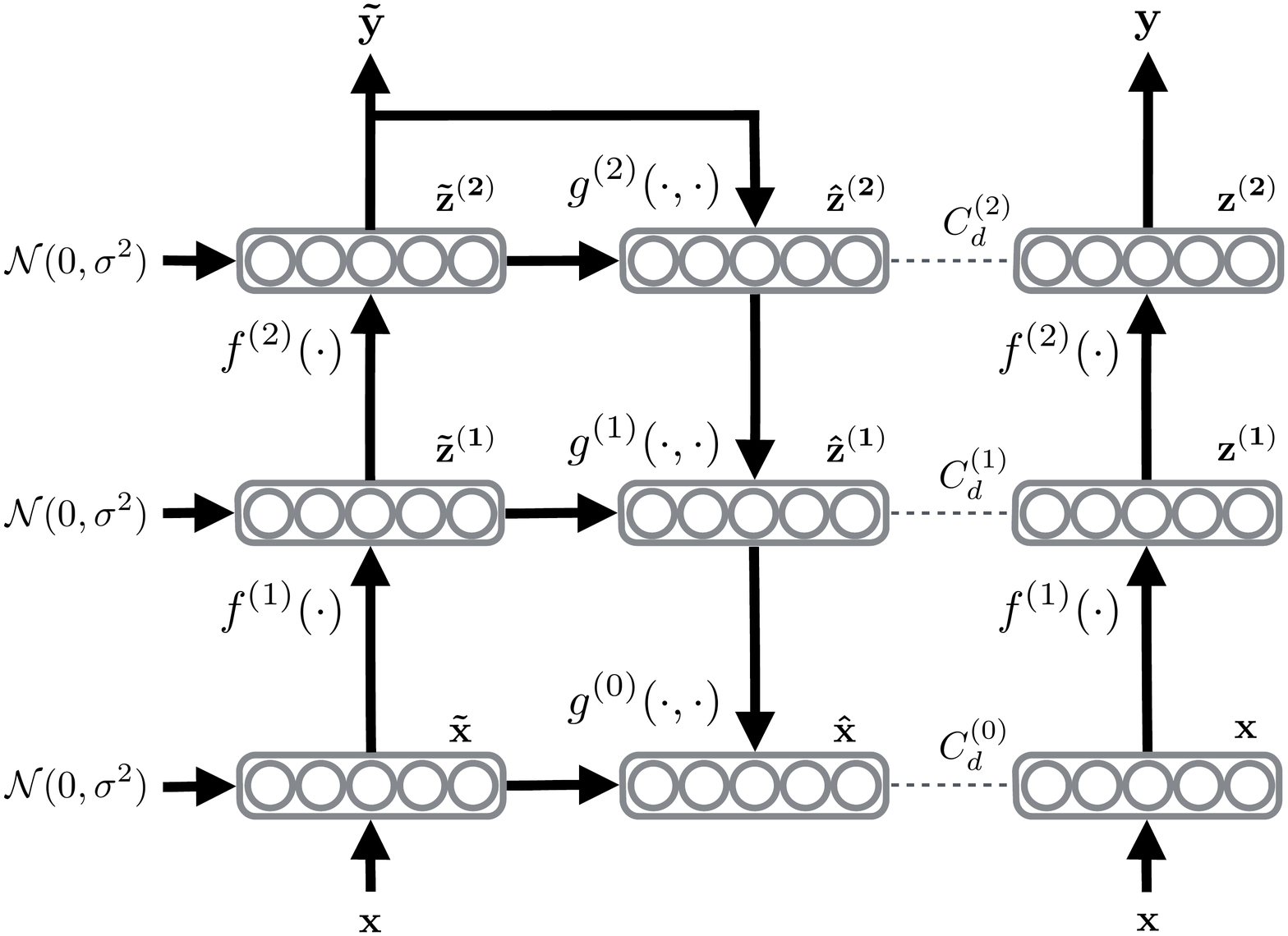}
\caption{A conceptual illustration of the Ladder network when $L=2$.
The feedforward path ($\x \to \z 1 \to \z 2 \to \y$) shares the mappings $\f l$ with
the corrupted feedforward path, or encoder ($\x \to \tz 1 \to \tz 2 \to \tilde{\y}$).
The decoder ($\tz l \to \hz l \to \hat{\x}$) consists of the denoising functions
$\g l$ and has cost functions
$C^{(l)}_d$ on each layer trying to minimize the difference between $\hz l $ and $\z l$.
The output $\tilde{\y}$ of the encoder can also be trained to match available labels $t(n)$.
}
\label{fig:ladder}
\end{center}
\end{figure}
% END ARXIV
}{
% BEGIN NIPS
\begin{figure}[tbp]
\begin{center}
  \includegraphics[width=0.49\columnwidth]{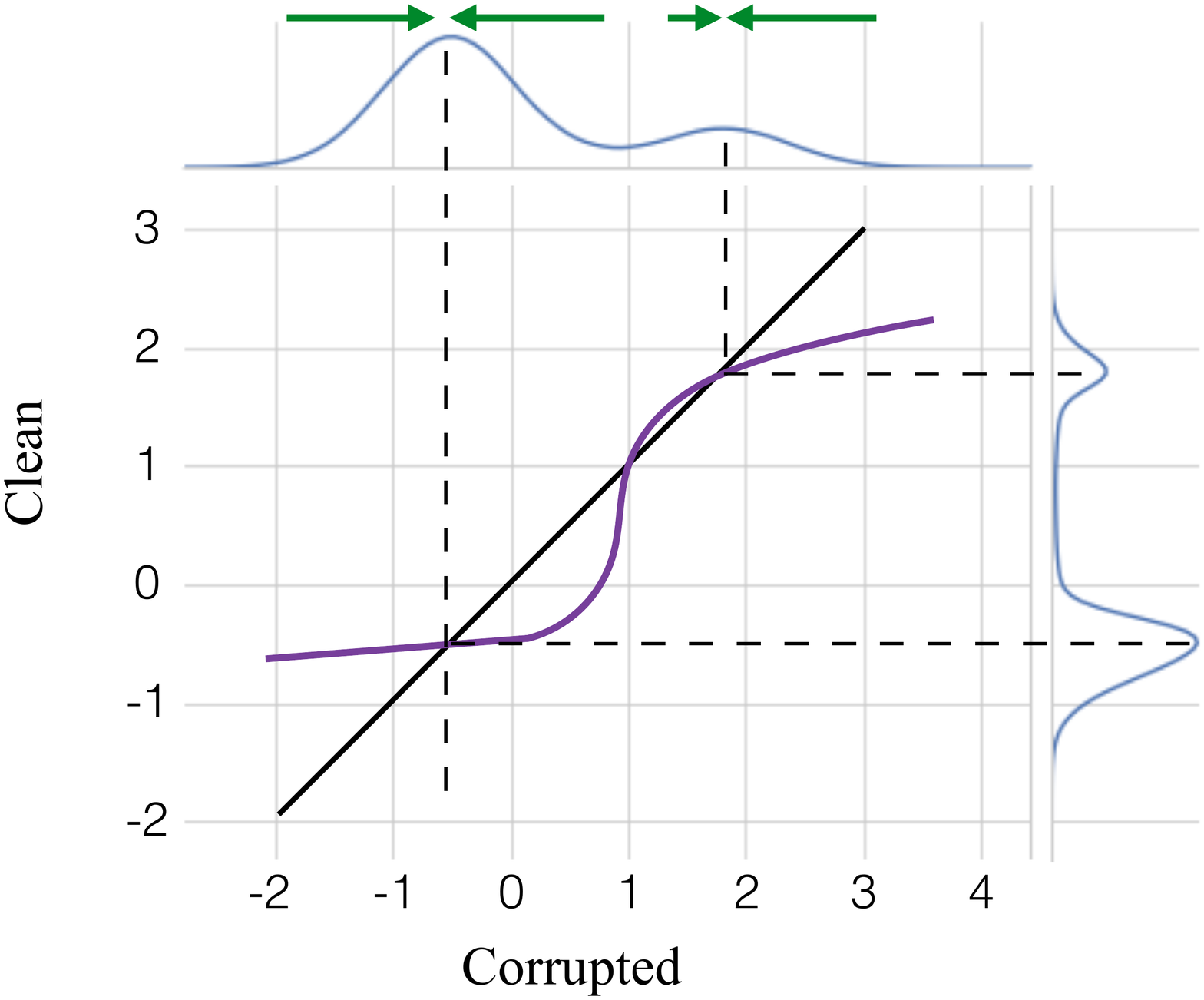}
  \includegraphics[width=0.49\columnwidth]{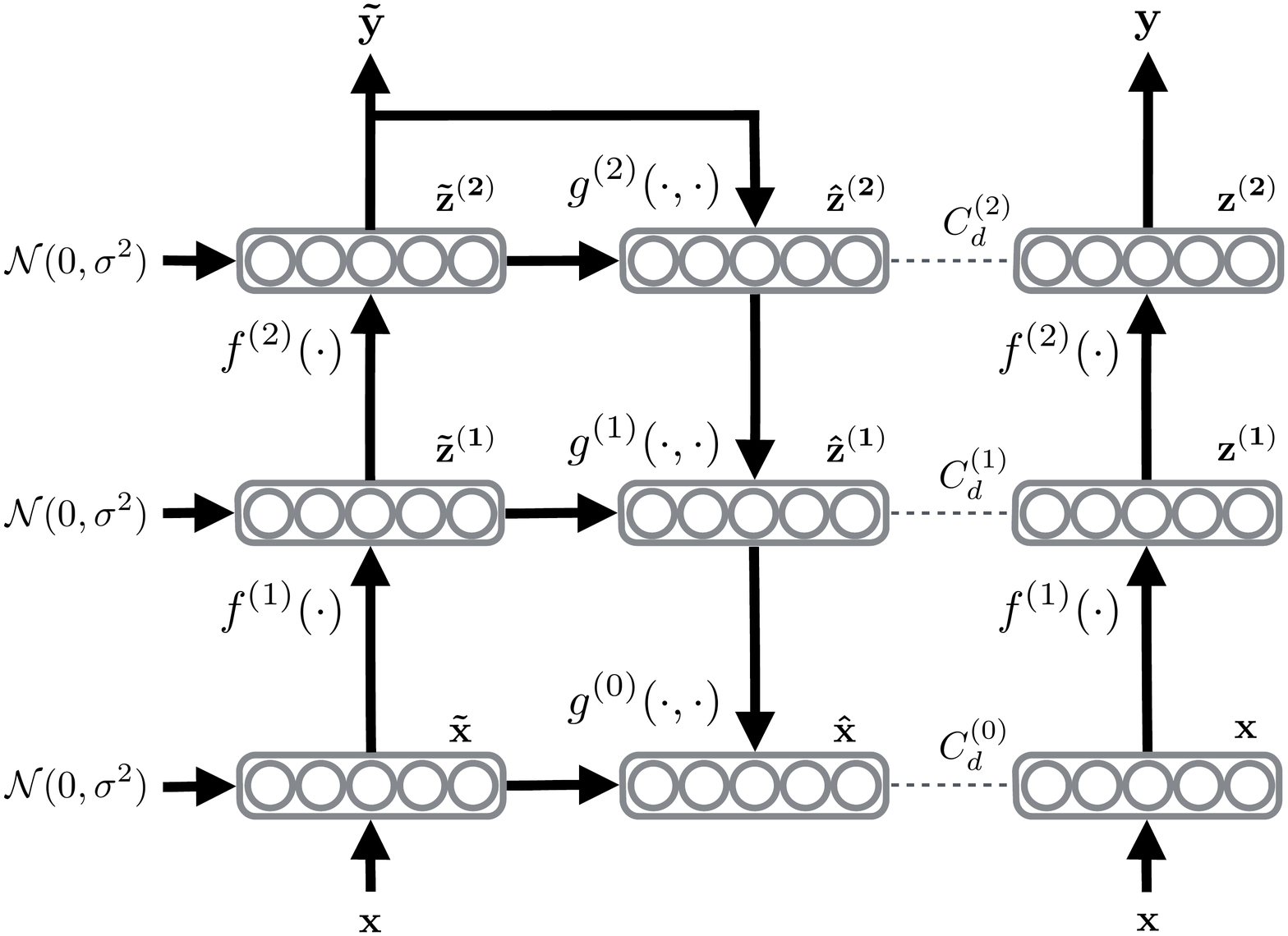}
\caption{\textbf{Left}: A depiction of an optimal denoising function for a bimodal distribution. The
input for the function is the corrupted value (x-axis) and the target is the clean
value (y-axis). The denoising function moves values towards higher probabilities as
show by the green arrows. \textbf{Right}: A conceptual illustration of the Ladder network when $L=2$.
The feedforward path ($\x \to \z 1 \to \z 2 \to \y$) shares the mappings $\f l$ with
the corrupted feedforward path, or encoder ($\x \to \tz 1 \to \tz 2 \to \tilde{\y}$).
The decoder ($\tz l \to \hz l \to \hat{\x}$) consists of denoising functions
$\g l$ and has cost functions
$C^{(l)}_d$ on each layer trying to minimize the difference between $\hz l $ and $\z l$.
The output $\tilde{\y}$ of the encoder can also be trained to match available labels $t(n)$.}
\label{fig:denois}
\label{fig:ladder}
\end{center}
\end{figure}

}

% % % % % % % % % % % % % % 
%\section{Algorithm for the feedforward pass of Ladder}
% % % % % % % % % % % % % % 

\begin{algorithm}
\caption{Calculation of the output $\vy$ and cost function $C$ of the Ladder network}
\label{alg:ff}
\begin{multicols}{2}
\begin{algorithmic}
\REQUIRE{$\x(n)$}

\# Corrupted encoder and classifier
\STATE $\tildeh 0 \leftarrow \tz 0 \leftarrow \x(n) + \mathtt{noise}$
\FOR{l = 1 \TO L}
	\STATE $\tz l \leftarrow \mathtt{batchnorm}(\W l \tildeh {l-1}) + \mathtt{noise}$
	\STATE $\tildeh l \leftarrow
   \mathtt{activation}(\boldsymbol\gamma\l \odot (\tz l + \boldsymbol\beta\l))$
\ENDFOR
\STATE  $P(\tilde{\y} \mid \x) \leftarrow \tildeh L$
\STATE
\# Clean encoder (for denoising targets)

\STATE $\h 0 \leftarrow \z 0 \leftarrow \x(n)$
\FOR{l = 1 \TO L}
    \STATE $\z l_{\mathrm{pre}} \leftarrow \W l \h {l-1}$
    \STATE ${\boldsymbol\mu}^{(l)} \leftarrow \mathtt{batchmean}(\z l_{\mathrm{pre}})$
    \STATE ${\boldsymbol\sigma}^{(l)} \leftarrow \mathtt{batchstd}(\z l_{\mathrm{pre}})$
    \STATE $\z l \leftarrow \mathtt{batchnorm} (\z l_{\mathrm{pre}})$
	\STATE $\h l \leftarrow \mathtt{activation}(\boldsymbol\gamma\l\odot(\z l + \boldsymbol\beta\l))$
\ENDFOR

	\columnbreak

	\# Final classification:
   \STATE $P(\y \mid \x) \leftarrow \h L$
   \STATE
	\# Decoder and denoising
    \FOR{l = L \TO 0}
    	\IF{l = L}
        	\STATE $\u L \leftarrow \mathtt{batchnorm} (\tildeh L)$
        \ELSE
		    \STATE $\u l \leftarrow \mathtt{batchnorm} (\V {l+1} \hz {l+1})$
		\ENDIF
		\STATE $\forall i: \hat{z}_i^{(l)} \leftarrow g(\tilde{z}_i^{(l)},u_i^{(l)})$
	    \s{\# Eq.\ (\ref{eq:g_func})}{}
		\STATE $\forall i: \hat{z}_{i,\mathrm{BN}}^{(l)} \leftarrow \frac{\hat{z}_i^{(l)} - \mu_i^{(l)}}{\sigma_i^{(l)}}$
    \ENDFOR
    
	\# Cost function $C$ for training:
   \STATE $\mathrm{C} \leftarrow 0$
   \IF{$t(n)$}
        \STATE $\mathrm{C} \leftarrow -\log P(\tilde{\y}=t(n) \mid \x(n))$
	\ENDIF
	\STATE $\mathrm{C} \leftarrow \mathrm{C} + \sum_{l=0}^L \lambda_l \left\Vert \z l - \hz l_{\mathrm{BN}} \right\Vert^2$ \s{\# Eq.\ (\ref{eq:cost})}{}
\end{algorithmic}
\end{multicols}
\end{algorithm}

One way to picture the Ladder network is to consider it as a collection of
nested denoising autoencoders which share parts of the
denoising machinery with each other. From the viewpoint of the
autoencoder on layer $l$, the representations on the higher layers
can be treated
as hidden neurons. In other words, there is no particular reason why
$\hz {l+i}$ as produced by the decoder should resemble the corresponding
representations $\z {l+i}$ as produced by the encoder. It is only the
cost function $C^{(l+i)}_d$ that ties these together and forces the
inference to proceed in reverse order in the decoder. This sharing
helps a deep denoising autoencoder to learn the denoising process as
it splits the
task into meaningful sub-tasks of denoising intermediate representations.

% % % % % % % % % % % % % % 
\section{Implementation of the Model}
% % % % % % % % % % % % % %
\label{sec:implementation}

\s{
The steps involved in implementing the Ladder network (Section~\ref{sec:howto}) are
typically as follows: 1) take a feedforward model which serves supervised
learning as the encoder (Section~\ref{sec:baseline}); 2) add a decoder
which can invert the mappings on each layer of the encoder
and supports unsupervised learning (Section~\ref{sec:model.unsuper});
and 3) train the whole Ladder network by
minimizing the sum of all the cost function terms.

In this section, we will go through these steps in detail for a fully connected
MLP network and briefly outline the modifications required for convolutional
networks, both of which are used in our experiments (Section~\ref{sec:experiments}).

% % % % % % % % % % % % % % 
\subsection{General Steps for Implementing the Ladder Network}
% % % % % % % % % % % % % % 
\label{sec:howto}

Consider training a classifier,%
\footnote{Here we only consider the case where the output $t(n)$ is a class label
but it is trivial to apply the same approach to other regression tasks.},
or a mapping from input $\vx$ to output 
$y$ with targets $t$, 
from a training set of pairs $\{\vx(n), t(n) \mid 1 \leq n \leq N\}$.
Semi-supervised learning \citep{chapelle2006semi} studies how 
auxiliary unlabeled data $\{\vx(n) \mid N+1 \leq n \leq M \}$
can help in training a classifier. It is 
often the case that labeled data are scarce whereas unlabeled data are 
plentiful, that is $N\ll M$.
%% TODO (Tapani): Eri notaatio kuin Related Work -osiossa.

The Ladder network can improve results even without auxiliary unlabeled
data but the original motivation was to make it possible to take
well-performing feedforward classifiers and augment them with an
auxiliary decoder as follows:
\begin{enumerate}
\item Train any standard feedforward neural network. The network type is not limited to standard MLPs,
but the approach can be applied, for example, to convolutional or recurrent networks.
This will be the encoder part of the Ladder network.
\item For each layer, analyze the conditional distribution of representations given the layer above,
$p(\z l \mid \z {l+1})$. The observed distributions could resemble for example Gaussian distributions where
the mean and variance depend on the values $\z {l+1}$, bimodal distributions where the relative probability
masses of the modes depend on the values $\z {l+1}$, and so on.

\item Define a function $\hz l = g(\tz l, \hz {l+1})$ which can
  approximate the optimal denoising function for the family of observed distributions. 
The function $g$ is therefore expected to form a reconstruction $\hz l$ that resembles the clean
$\z l$ given the corrupted $\tz l $ and the higher-level reconstruction $\hz {l+1}$ .
\item Train the whole network in a fully-labeled or semi-supervised setting using
  standard optimization techniques such as stochastic gradient descent.
\end{enumerate}
% END OF ARXIV VERSION
}{
% NIPS VERSION
\label{sec:variations}
We implement the Ladder network for fully connected MLP networks and for convolutional networks.
We used standard rectifier networks with batch normalization applied to each preactivation.
The feedforward pass of the full Ladder
network is listed in Algorithm~\ref{alg:ff}.

In the decoder, we parametrize the denoising function such that it supports denoising
of conditionally independent Gaussian latent variables, conditioned on the activations $\hz{l+1}$ of the layer above.
%The form of the denoising function $g$ is therefore 
The denoising function $g$ is therefore coupled into components
$\hat z^{(l)}_i = g_i(\tilde{z}^{(l)}_i, u_i^{(l)}) = \left(\tilde{z}^{(l)}_i - \mu_i (u_i^{(l)})\right) \upsilon_i (u_i^{(l)}) + \mu_i (u_i^{(l)}) \,
$
where $u_i^{(l)}$ propagates information from $\hz{l+1}$ by
%We choose the functional form of $\u l$ to be
%a mapping from $\hz {l+1}$, which is then batch normalized:
$
	\u l = \mathtt{batchnorm}(\V {l+1} \hz {l+1}) \, .
$
The functions $\mu_i (u_i^{(l)})$ and $\upsilon_i (u_i^{(l)})$ are modeled
as expressive nonlinearities:
$
\mu_i (u_i^{(l)}) = a_{1,i}^{(l)} \mathtt{sigmoid}(a_{2,i}^{(l)} u_i^{(l)} + a_{3,i}^{(l)}) + a_{4,i}^{(l)} u_i^{(l)} + a_{5,i}^{(l)}
$,
with the form of the nonlinearity similar for $\upsilon_i (u_i^{(l)})$.
The decoder has thus 10 unit-wise parameters $a$, compared to the two parameters ($\gamma$ and $\beta$ \citep{Ioffe15icml}) in the encoder.

% The cost function for the unsupervised path is the mean squared reconstruction
% error per neuron. However, we implicitly use the projections $\z l_{\mathrm{pre}}$
% as the target for denoising and scale the cost function in such a way that
% the term appearing in the error term is the batch normalized $\z l$ instead.
% For the moment, let us see how that works for a scalar case:
% \begin{eqnarray*}
% 	\frac{1}{\sigma^2} \left\Vert z_{\mathrm{pre}} - \hat z \right\Vert ^2 & = &
%     \left\Vert \frac{z_{\mathrm{pre}} - \mu}{\sigma} - \frac{\hat{z} - \mu}{\sigma}\right\Vert^2 =
%     \left\Vert z - \hat{z}_{\mathrm{BN}}\right\Vert^2 \\
%     z & = & \mathrm{N_B}(z_{\mathrm{pre}}) = \frac{z_{\mathrm{pre}} - \mu}{\sigma} \\
%     \hat{z}_{\mathrm{BN}} & = & \frac{\hat{z} - \mu}{\sigma} \, ,
% \end{eqnarray*}

% where $\mu$ and $\sigma$ are the batch mean and batch std of $z_{\mathrm{pre}}$,
% respectively, that were used in batch normalizing $z_{\mathrm{pre}}$ into $z$. 
% The unsupervised denoising cost function $C_\mathrm{d}$ is thus
% \begin{equation}
% C_\mathrm{d} = \sum_{l=0}^L \lambda_l C_\mathrm{d}^{(l)}
% = \sum_{l=0}^L \frac {\lambda_l} {Nm_l} \sum_{n=1}^N \left\Vert \z l(n) - \hz l_{\mathrm{BN}}(n) \right\Vert^2,
% \label{eq:cost}
% \end{equation}
% where $m_l$ is the layer's width, N the number of training samples, and the hyperparameter
% $\lambda_l$ a layer-wise multiplier determining the importance of the denoising cost.

It is worth noting that a simple special case of the decoder is a model where
$\lambda_l = 0$ when $l < L$. This corresponds to a denoising cost only on the top layer
and means that most of the decoder can be omitted. This model, which we call the $\Gamma$-model
due to the shape of the graph, is useful as it can easily be plugged into any feedforward
network without decoder implementation.

Further implementation details of the model can be found in the supplementary material or Ref.~\cite{rasmus2015semi}.
}

\s{
% % % % % % % % % % % % % % 
\subsection{Fully Connected MLP as Encoder}
% % % % % % % % % % % % % % 
\label{sec:baseline}

As a starting point we use
a fully connected MLP network with rectified linear units.
We follow \citet{Ioffe15icml} and apply batch normalization to
each preactivation including the topmost layer in the $L$-layer network.
This serves two purposes. First, it improves convergence as a result of
reduced covariate shift as originally proposed by \citet{Ioffe15icml}.
Second, as explained in Section~\ref{sec:derivation},
DSS-type cost functions for all but
the input layer require some type of normalization to prevent the denoising
cost from encouraging the trivial solution where the encoder outputs just
constant values as these are the easiest to denoise. Batch normalization
conveniently serves this purpose, too.

Formally, batch normalization for the layers $l=1\dots L$ is implemented as
$$\z l = \mathrm{N_B}(\W l \h {l-1})$$
$$\h l = \phi\big(\boldsymbol{\gamma}^{(l)}(\z l + \boldsymbol{\beta}^{(l)})\big),$$
where $\h 0 = \x$, $\mathrm{N_B}$ is a component-wise batch normalization $\mathrm{N_B}(x_i) =
(x_i - \hat \mu_{x_i})/{\hat \sigma_{x_i}}$,
where $\hat \mu_{x_i}$ and $\hat \sigma_{x_i}$ are estimates calculated from
the minibatch, $\boldsymbol\gamma^{(l)}$ and $\boldsymbol\beta^{(l)}$ are trainable  
parameters, and $\phi(\cdot)$ is the activation function such as the
rectified linear unit (ReLU) for which $\phi(\cdot) =\max (0,\cdot)$.
For outputs $\y = \h L$ we always use the softmax activation. For some
activation functions the scaling parameter $\boldsymbol\beta^{(l)}$
or the bias $\boldsymbol\gamma^{(l)}$ are redundant and we only apply
them in non-redundant cases. For example, the rectified linear unit
does not need scaling, the linear activation function needs neither scaling
nor bias, but softmax requires both.

As explained in Section~\ref{sec:derivation} and shown in
Figure~\ref{fig:ladder}, the Ladder network requires two forward passes,
one clean and one corrupted,
which produce clean $\z l$ and $\h l$ and corrupted $\tz l$ and
$\tildeh l$, respectively. We implemented corruption by adding
isotropic Gaussian noise $\vect{n}$ to inputs and after each batch
normalization:
\begin{eqnarray*}
\tilde \x & = & \tildeh 0 = \x + \vect{n}^{(0)} \\
\tz l_{\mathrm{pre}} & = & \W l \tildeh {l-1} \\
\tz l & = & \mathrm{N_B}(\tz l_{\mathrm{pre}}) + \vect{n}^{(l)} \\
\tildeh l & = & \phi\big(\boldsymbol{\gamma}^{(l)}(\tz l + \boldsymbol{\beta}^{(l)})\big).
\end{eqnarray*}
Note that we collect the value $\tz l_{\mathrm{pre}}$ here because it will be
needed in the decoder cost function in Section~\ref{sec:model.unsuper}.

The supervised cost $C_\mathrm{c}$ is the average negative log probability of the
noisy output $\tilde \y$ matching the target $t(n)$ given the inputs $\mathbf{x}(n)$
$$C_\mathrm{c} = - \frac 1 N \sum_{n=1}^N \log P(\tilde \y=t(n)\mid 
\mathbf{x}(n)).$$
In other words, we also use the noise to regularize supervised learning.

We saw networks with this structure reach close to state-of-the-art results
in purely supervised learning (see e.g. Table~\ref{tab:semisup_results}),
which makes them good starting points for improvement via semi-supervised
learning by adding an auxiliary unsupervised task.

% % % % % % % % % % % % % % 
\subsection{Decoder for Unsupervised Learning}
% % % % % % % % % % % % % % 
\label{sec:model.unsuper}

When designing a suitable decoder to support unsupervised learning, we had to
make a choice as to what kinds of distributions of the latent variables the decoder would optimally
be able to denoise. We ultimately ended up choosing a parametrization that
supports the optimal denoising of Gaussian latent variables. We also experimented
with alternative denoising functions, more details of which can be found in
Appendix~\ref{sec:comparison}. Further analysis of different denoising functions
was recently published by \citet{deconstructingladder}.

In order to derive the chosen parametrization and justify why it supports Gaussian
latent variables, let us begin with the assumption that the noisy value of one latent variable
$\tilde z$ that we want to denoise has the form $\tilde z = z + n$, where
$z$ is the clean latent variable value that has a Gaussian distribution with variance $\sigma_z^2$,
and $n$ is the Gaussian noise with variance $\sigma_n^2$.

We now want to estimate $\hat z$, a denoised version of $\tilde z$, so that the estimate
minimizes the squared error of the difference to the clean latent variable values $z$. It can
be shown that the functional form of $\hat z = g(\tilde z)$ has to be linear
in order to minimize the denoising cost, with the assumption being that both the noise and the
latent variable have a Gaussian distribution \citep[][Section 4.1]{valpola2015ladder}.
Specifically, the result will be a weighted
sum of the corrupted $\tilde z$ and a prior $\mu$. The weight $\upsilon$
of the corrupted $\tilde z$ will be a function of the variance of $z$ and $n$
according to:
$$
\upsilon = \frac{\sigma_{z}^2}{\sigma_{z}^2 + \sigma_n^2}
$$

The denoising function will therefore have the form:
\begin{equation}
\label{eq:simple_g}
\hat z = g(\hat z) = \upsilon * \tilde{z} + (1 - \upsilon) * \mu = (\tilde{z} - \mu) * \upsilon + \mu
\end{equation}

We could let $\upsilon$ and $\mu$ be trainable parameters of the model, where the 
model would learn some estimate of the optimal weighting $\upsilon$ and prior $\mu$.
The problem with this formulation is that it only supports the optimal denoising of latent
variables with a Gaussian distribution, as the function $g$ is linear wrt. $\tilde z$.

We relax this assumption by making the model only require the distribution of $z$ of a layer
to be Gaussian conditional on the values of the latent variables of the layer above. In a similar vein,
in a layer of multiple latent variables we can assume that the latent variables are independent
conditional on the latent variables of the layer above. The distribution of the latent variables
$\z l$ is therefore assumed to follow the distribution
$$p(\z l \mid \z {l+1}) = \prod_i p(z_i^{(l)} \mid \z {l+1})$$
where $p(z_i^{(l)} \mid \z {l+1})$
are conditionally independent Gaussian distributions.

% The reason this formulation only supports
% optimal denoising of independent distributions is that the lateral connections
% are diagonal, i.e. $\hat z^{(l)}_i$ only depends on $\tilde{z}^{(l)}_i$ and not the full $\tz l$.

% One way to relax the assumption of independent and Gaussian distributions is to make the
% coefficients $\vv$ and $\vmu$ depend on the above layer values $\hz {l+1}$. In that case we assume
% the values $\z l$ are still independent Gaussians, however conditioned on $\z {l+1}$.
One interpretation of this formulation is that we are modeling the distribution of $\z l$ as a mixture of
Gaussians with diagonal covariance matrices, where the value of the above layer
$\z {l+1}$ modulates the form of the Gaussian that $\z l$ is distributed as.
In practice, we will implement the dependence of $\vv$ and $\vmu$ on $\hz {l+1}$
with a batch normalized projection from $\hz {l+1}$ followed by an expressive nonlinearity with
trainable parameters.
The final formulation of the denoising function is therefore

\begin{equation}
\hat z^{(l)}_i = g_i(\tilde{z}^{(l)}_i, u_i^{(l)}) = \left(\tilde{z}^{(l)}_i - \mu_i (u_i^{(l)})\right) \upsilon_i (u_i^{(l)}) + \mu_i (u_i^{(l)}) \,
\label{eq:g_func}
\end{equation}
where $u_i^{(l)}$ propagates information from $\hz{l+1}$ by a batch normalized projection:
$$
	\u l = \mathrm{N_B}(\V {l+1} \hz {l+1}) \, ,
$$
where the matrix $\V l$ has the same dimension as the transpose of $\W l$ on the
encoder side. The projection vector $\u l$ therefore has the same dimensionality as $\z l$.
Furthermore, the functions $\mu_i (u_i^{(l)})$ and $\upsilon_i (u_i^{(l)})$ are modeled
as expressive nonlinearities:
$$
\mu_i (u_i^{(l)}) = a_{1,i}^{(l)} \mathtt{sigmoid}(a_{2,i}^{(l)} u_i^{(l)} + a_{3,i}^{(l)}) + a_{4,i}^{(l)} u_i^{(l)} + a_{5,i}^{(l)}
$$
$$
\upsilon_i (u_i^{(l)}) = a_{6,i}^{(l)} \mathtt{sigmoid}(a_{7,i}^{(l)} u_i^{(l)} + a_{8,i}^{(l)}) + a_{9,i}^{(l)} u_i^{(l)} + a_{10,i}^{(l)},
$$
where $a_{1,i}^{(l)} \dots a_{10,i}^{(l)}$ are the trainable parameters of the nonlinearity
for each neuron $i$ in each layer $l$. It is worth noting that in this parametrization,
each denoised value $\hat z^{(l)}_i$ only depends on $\tilde{z}^{(l)}_i$ and not the full $\tz l$.
This means that the model can only optimally denoise conditionally independent distributions.
While this nonlinearity makes the number of parameters in the decoder slightly higher than in the
encoder, the difference is insignificant as most of the parameters are in the vertical
projection mappings $\W l$ and $\V l$, which have the same dimensions (apart from
transposition). Note the slight abuse of the notation here since $g_i^{(l)}$ is now a function of
the scalars $\tilde z^{(l)}_i$ and $u^{(l)}_i$ rather than the full
vectors $\tz l$ and $\hz {l+1}$.
Given $\u l$, this parametrization 
is linear with respect to $\tz l$, and
both the slope and the bias depend nonlinearly on $\u l$, as we hoped.

For the lowest layer, $\hat{\x} = \hz 0$ and $\tilde \x = \tz 0$ by definition,
and for the highest layer we chose $\u L = \tilde \y$. This allows the highest-layer
denoising function to utilize prior information about the classes being mutually
exclusive, which seems to improve convergence in cases where there are very few
labeled samples.

As a side note, if the values of $\z l$ are truly independently distributed Gaussians, there
is nothing left for the layer above, $\hz {l+1}$, to model. In that case,
a mixture of Gaussians is not needed to model $\z l$, but a diagonal Gaussian which
can be modeled with a linear denoising function with constant values for $\vupsilon$ and $\mu$ as in
Equation~\ref{eq:simple_g}, would suffice.
In this parametrization all correlations, nonlinearities,
and non-Gaussianities in the latent variables $\z l$ have to be represented by modulations
from the layers above for optimal denoising.
As the parametrization allows the distribution of $\z l$ to be
modulated by $\z {l+1}$ through $\u l$,
it encourages the decoder to find representations $\z l$ that have
high mutual information with $\z {l+1}$. This is crucial as it allows
supervised learning to have an indirect influence on the representations
learned by the unsupervised decoder: any abstractions selected by supervised
learning will bias the lower levels to find more representations which
carry information about the same abstractions.

The cost function for the unsupervised path is the mean squared reconstruction
error per neuron, but there is a slight twist which we found to be important.
Batch normalization has useful properties, as noted in Section~\ref{sec:baseline},
but it also introduces noise which affects both the clean and corrupted
encoder pass. This noise is highly correlated between $\z l$ and
$\tz l$ because the noise derives from the statistics of the samples that
happen to be in the same minibatch. This highly correlated noise in $\z l$
and $\tz l$ biases the denoising functions to be simple copies%
\footnote{The whole point of using \textit{denoising} autoencoders rather
than regular autoencoders is to prevent skip connections from short-circuiting
the decoder and force the decoder to learn meaningful abstractions which
help in denoising.} $\hz l \approx \tz l$.

The solution we found was to implicitly use the projections $\z l_{\mathrm{pre}}$
as the target for denoising and scale the cost function in such a way that
the term appearing in the error term is the batch normalized $\z l$ instead.
For the moment, let us see how that works for a scalar case:
\begin{eqnarray*}
	\frac{1}{\sigma^2} \left\Vert z_{\mathrm{pre}} - \hat z \right\Vert ^2 & = &
    \left\Vert \frac{z_{\mathrm{pre}} - \mu}{\sigma} - \frac{\hat{z} - \mu}{\sigma}\right\Vert^2 =
    \left\Vert z - \hat{z}_{\mathrm{BN}}\right\Vert^2 \\
    z & = & \mathrm{N_B}(z_{\mathrm{pre}}) = \frac{z_{\mathrm{pre}} - \mu}{\sigma} \\
    \hat{z}_{\mathrm{BN}} & = & \frac{\hat{z} - \mu}{\sigma} \, ,
\end{eqnarray*}
% TODO: Saatiin tällainen kommentti, joka on musta oikein. Koodikin normalisoi z-tilde-pre:n 
% statistiikan avulla!
%
%- On page 6, end of section 3.3, z-hat_BN is normalized by batch mean and batch std of z_pre (clean Z). 
% But on page 4 "Algorithm 1", z-hat_BN is normalized instead by batch mean and std of z-tilde_pre
%(corrupted Z). I wonder what difference does this make?

where $\mu$ and $\sigma$ are the batch mean and batch std of $z_{\mathrm{pre}}$,
respectively, that were used in batch normalizing $z_{\mathrm{pre}}$ into $z$. 
The unsupervised denoising cost function $C_\mathrm{d}$ is thus
\begin{equation}
C_\mathrm{d} = \sum_{l=0}^L \lambda_l C_\mathrm{d}^{(l)}
= \sum_{l=0}^L \frac {\lambda_l} {Nm_l} \sum_{n=1}^N \left\Vert \z l(n) - \hz l_{\mathrm{BN}}(n) \right\Vert^2,
\label{eq:cost}
\end{equation}
where $m_l$ is the layer's width, N the number of training samples, and the hyperparameter
$\lambda_l$ a layer-wise multiplier determining the importance of the denoising cost.

The model parameters $\W l,\boldsymbol{\gamma}^{(l)},\boldsymbol{\beta}^{(l)},\V l,\mathbf{a}_i^{(l)}, b_i^{(l)}, and \mathbf{c}_i^{(l)}$
can be trained simply by using the backpropagation algorithm to optimize
the total cost $C = C_\mathrm{c} + C_\mathrm{d}$. The feedforward pass of the full Ladder
network is listed in Algorithm~\ref{alg:ff}. Classification results are
read from the $\y$ in the clean feedforward path.

\subsection{Variations}
\label{sec:variations}

Section \ref{sec:model.unsuper} detailed how to build a decoder for
the Ladder network to match the fully connected encoder described in
Section~\ref{sec:baseline}. It is easy to extend the same approach to other
encoders, for instance, convolutional neural networks (CNN). For the
decoder of fully connected networks we used vertical mappings whose
shape is a transpose of the encoder mapping. The same treatment works
for the convolution operations: in the networks we have tested in this
paper, the decoder has convolutions whose parametrization mirrors the
encoder and effectively just reverses the flow of information. As the
idea of convolution is to reduce the number of parameters by weight
sharing, we applied this to the parameters of the denoising function
g, too.

Many convolutional networks use pooling operations with stride; that
is, they downsample the spatial feature maps. The decoder needs to
compensate for this with a corresponding upsampling. There are several
alternative ways to implement this and in this paper we chose the
following options: 1) on the encoder side, pooling operations are
treated as separate layers with their own batch normalization and
linear activation function, and 2) the downsampling of the pooling on
the encoder side is compensated for by upsampling with copying on the
decoder side. This provides multiple targets for the decoder to match,
helping the decoder to recover the information lost on the encoder
side.

It is worth noting that a simple special case of the decoder is a model where
$\lambda_l = 0$ when $l < L$. This corresponds to a denoising cost only on the top layer
and means that most of the decoder can be omitted. This model, which we call the $\Gamma$-model
because of the shape of the graph, is useful as it can easily be plugged into any feedforward
network without decoder implementation. In addition, the $\Gamma$-model is the same
for MLPs and convolutional neural networks. The encoder in the $\Gamma$-model
still includes both the clean and the corrupted paths as in the full ladder.
% END OF ARXIV VERSION
}{
% NIPS VERSION
}

% % % % % % % % % % % % % % 
\section{Experiments}
% % % % % % % % % % % % % % 
\label{sec:experiments}

\s{
With the experiments with the MNIST and CIFAR-10 datasets, we wanted to compare our method to other semi-supervised methods but
also show that we can attach the decoder both to a fully connected MLP network and to a convolutional neural network, both of which were described
in Section~\ref{sec:implementation}.
We also wanted to compare the performance of the simpler $\Gamma$-model
(Sec.~\ref{sec:variations}) to the full Ladder network
and experimented with only having a cost function on the input layer.
With CIFAR-10, we only tested the $\Gamma$-model.
}{
We ran experiments both with the MNIST and CIFAR-10 datasets, where we attached the decoder both
to fully connected MLP networks and to convolutional neural networks. We also compared
the performance of the simpler $\Gamma$-model
(Sec.~\ref{sec:variations}) to the full Ladder network.
}

\s{
We also measured the performance of the supervised baseline models
which only included the encoder and the supervised cost function. In
all cases where we compared these directly with Ladder networks, we
did our best to optimize the hyperparameters and regularization of the
baseline supervised learning models so that any improvements could not
be explained, for example, by the lack of suitable regularization
which would then have been provided by the denoising costs.}{}

\s{
With convolutional networks, our focus was exclusively on
semi-supervised learning. The supervised baselines for all labels only
intend to show that the performance of the selected network
architectures is in line with the ones reported in the literature.
We make claims neither about the optimality nor the statistical
significance of these baseline results.
}{
With convolutional networks, our focus was exclusively on
semi-supervised learning.
We make claims neither about the optimality nor the statistical
significance of the supervised baseline results.
}

\s{
We used the Adam optimization algorithm \citep{kingma2015adam} for the weight
updates. The learning rate was 0.002 for the first part of the learning,
followed by an annealing phase during which the learning rate was
linearly reduced to zero. The minibatch size was 100.
The source code for all the experiments is available at
\url{https://github.com/arasmus/ladder} unless explicitly noted in the text.
}{
We used the Adam optimization algorithm \citep{kingma2015adam}. The initial
learning rate was 0.002 and it was decreased linearly to zero
during a final annealing phase. The minibatch size was 100.
The source code for all the experiments is available at
\url{https://github.com/arasmus/ladder}.
}

\subsection{MNIST dataset}

\s{
For evaluating semi-supervised learning, we used
the standard 10,000 test samples as a held-out test set and
randomly split the standard 60,000 training samples
into a 10,000-sample validation set and used $M=50,000$ samples as
the training set. From the training
set, we randomly chose $N=100$, $1000$, or all labels for the supervised cost.%
\footnote{In all the experiments, we were careful not to optimize any
parameters, hyperparameters, or model choices on the basis of the results
on the held-out test samples. As is customary, we used 10,000
labeled validation samples even for those settings where we only
used 100 labeled samples for training. Obviously, this is not something
that could be done in a real case with just 100 labeled samples.
However, MNIST classification is such an easy task, even in the
permutation invariant case, that 100 labeled samples
there correspond to a far greater number of labeled samples in many
other datasets.}
All the samples were used
for the decoder, which does not need the labels. The validation set was used for evaluating the model structure and
hyperparameters. We also balanced the classes to ensure that no particular class was
over-represented. We repeated each training 10 times, varying the random seed that was used
for the splits.

After optimizing the hyperparameters, we performed the final test runs using all
the $M=60,000$ training samples with 10 different random initializations of the weight matrices and data splits.
We trained all the models for 100 epochs followed by 50 epochs of annealing.
With minibatch size of 100, this amounts to 75,000 weight updates for the validation
runs and 90,000 for the final test runs.
}{
For evaluating semi-supervised learning, we 
randomly split the 60,000 training samples
into 10,000-sample validation set and used $M=50,000$ samples as
the training set. From the training
set, we randomly chose $N=100$, $1000$, or all labels for the supervised cost.%
\footnote{In all the experiments, we were careful not to optimize any
parameters, hyperparameters, or model choices on the basis of the results
on the held-out test samples. As is customary, we used 10,000
labeled validation samples even for those settings where we only
used 100 labeled samples for training. Obviously, this is not something
that could be done in a real case with just 100 labeled samples.
However, MNIST classification is such an easy task, even in the
permutation invariant case, that 100 labeled samples
there correspond to a far greater number of labeled samples in many
other datasets.}
All the samples were used
for the decoder, which does not need the labels. The validation set was used for evaluating the model structure and
hyperparameters. We also balanced the classes to ensure that no particular class was
over-represented. We repeated the training 10 times varying the random seed
for the splits.

After optimizing the hyperparameters, we performed the final test runs using all
the $M=60,000$ training samples with 10 different random initializations of the weight matrices and data splits.
We trained all the models for 100 epochs followed by 50 epochs of annealing.
}

\subsubsection{Fully connected MLP}

A useful test for general learning algorithms is the permutation
invariant MNIST classification task.
\s{Permutation invariance means that
the results need to be invariant with respect to permutation of the
elements of the input vector. In other words, one is not allowed to
use prior information about the spatial arrangement of the input
pixels. This excludes, among others, convolutional networks and
geometric distortions of the input images.

We chose the layer sizes of the baseline model to be
784-1000-500-250-250-250-10. The network is deep 
enough to demonstrate the scalability of the method but does not yet represent overkill for MNIST.
}{
We chose the layer sizes of the baseline model to be
784-1000-500-250-250-250-10.} 

\s{
The hyperparameters we tuned for each model are the noise level that is added to the inputs and
to each layer, and the denoising cost multipliers $\lambda \l$. We also ran the
supervised baseline model with various noise levels.
For models with just one cost multiplier, we optimized them
with a search grid $\{\ldots$, 0.1, 0.2, 0.5, 1, 2, 5, 10, $\ldots$\}.
Ladder networks with a cost function on all their layers have a much larger
search space and we explored it much more sparsely.
For instance, the
optimal model we found for $N=100$ labels had $\lambda^{(0)}=1000$,
$\lambda^{(1)}=10$, and $\lambda^{(\geq 2)} = 0.1$. A good value for the std
of the Gaussian corruption noise $\vect{n}^{(l)}$ was mostly $0.3$
but with $N=1000$ labels, $0.2$ was a better value.
For the complete set of selected denoising cost multipliers
and other hyperparameters, please refer to the code.
}{
The hyperparameters we tuned for each model are the noise level that is added to the inputs and
to each layer, and the denoising cost multipliers $\lambda \l$. We also ran the
supervised baseline model with various noise levels.
For models with just one cost multiplier, we optimized them
with a search grid $\{\ldots$, 0.1, 0.2, 0.5, 1, 2, 5, 10, $\ldots$\}.
Ladder networks with a cost function on all their layers have a much larger
search space and we explored it much more sparsely.
For the complete set of selected denoising cost multipliers
and other hyperparameters, please refer to the code.
}

\begin{table}[t]
 \caption{A collection of previously reported MNIST test errors in
 the permutation invariant setting followed by the results with
 the Ladder network. *~=~SVM. Standard deviation in parentheses.
 }
\begin{center}
 \begin{tabular}{llllll}
   Test error \% with \# of used labels & 100 & 1000 & All \\
   \hline
   Semi-sup. Embedding \citep{weston2012deep} & 16.86 & 5.73 & 1.5 \\
   Transductive SVM \citep[from][]{weston2012deep} & 16.81 & 5.38 & 1.40* \\
   MTC \citep{rifai2011manifold} & 12.03 & 3.64 & 0.81 \\ 
   Pseudo-label \citep{lee2013pseudo} & 10.49  & 3.46 & \\
   AtlasRBF \citep{pitelis2014semi} & 8.10 ($\pm$ 0.95) & 3.68 ($\pm$ 0.12) & 1.31 \\
   DGN \citep{kingma2014semi} & 3.33 ($\pm$ 0.14) & 2.40 ($\pm$ 0.02) & 0.96 \\
   DBM, Dropout \citep{srivastava2014dropout} & &  & 0.79 \\
   Adversarial \citep{goodfellow2015adver} &  & & 0.78 \\
   Virtual Adversarial \citep{miyato2015} & 2.12 & 1.32 & 0.64 ($\pm$ 0.03) \\
   \hline
   Baseline: MLP, BN, Gaussian noise & 21.74 ($\pm$ 1.77) & 5.70 ($\pm$ 0.20) &  0.80 ($\pm$ 0.03) \\
   $\Gamma$-model (Ladder with only top-level cost) & 3.06 ($\pm$ 1.44) & 1.53 ($\pm$ 0.10) & 0.78 ($\pm$ 0.03) \\
   Ladder, only bottom-level cost & 1.09 ($\pm 0.32$) &  0.90 ($\pm$ 0.05) & 0.59 ($\pm$ 0.03) \\
   Ladder, full  & {\bf{1.06}} ($\pm$ 0.37) &  {\bf{0.84}} ($\pm$ 0.08) & {\bf{0.57}} ($\pm$ 0.02)\\

\end{tabular}
 \end{center}
 % TODO: NIPS reviews: explain why some cells are empty
 % TODO: NIPS reviews: explain which comparison papers use "class balancing" if not all
 \label{tab:semisup_results}
\end{table}

The results presented in Table~\ref{tab:semisup_results} show that the
proposed method outperforms all the previously reported results.
% The improvement is most significant in the most difficult 100-label case
% where denoising targets on all layers provides the greatest benefit
% over having a denoising target only on the input layer. This suggests
% that the improvement can be attributed to efficient unsupervised
% learning on all the layers of the Ladder network.
Encouraged by the
good results, we also tested with $N=50$ labels and got a test error
of 1.62~\% ($\pm$ 0.65~\%).

% The simple $\Gamma$-model also performed surprisingly well,
% particularly for $N=1000$ labels.
% The denoising cost at the highest
% layer turns out to encourage distributions with two sharp peaks. While
% this clearly allows the model to utilize information in unlabeled
% samples, effectively self-labeling them, it also seems to suffer from
% confirmation bias, particularly with less labels. While the median
% error with $N=100$ labels is $2.61~\%$, the average is significantly
% worse due to some runs in which the model seems to get stuck with
% its inital misconception about the classes. The Ladder network with
% denoising targets on every layer converges much more reliably as can
% be seen from the low standard deviation of the results.

The simple $\Gamma$-model also performed surprisingly well, particularly
for $N=1000$ labels. With $N=100$ labels, all the models sometimes failed to converge
properly. With bottom level or full costs in Ladder, around 5~\% of runs result in a test error
of over 2~\%. In order to be able to estimate the average test error reliably in the
presence of such random outliers, we ran 40 instead of 10 test runs with random
initializations. % TODO: Check the numbers

\subsubsection{Convolutional networks}

\s{
We tested two convolutional networks for the general MNIST
classification task but omitted data augmentation such as geometric
distortions. We focused on the 100-label case since with more labels
the results were already so good even in the more difficult
permutation invariant task.

The first network was a straightforward extension of the
fully connected network tested in the permutation invariant case.  We
turned the first fully connected layer into a convolution with
26-by-26 filters, resulting in a 3-by-3 spatial map of 1000 features.
Each of the nine spatial locations was processed independently by a
network with the same structure as in the previous section, finally
resulting in a 3-by-3 spatial map of 10 features. These were pooled
with a global mean-pooling layer. Essentially we thus convolved the
image with the complete fully connected network.
Depooling on the
topmost layer and deconvolutions on the layers below were implemented
as described in Section~\ref{sec:variations}. Since the internal
structure of each of the nine almost independent processing paths was the
same as in the permutation invariant task, we used the same
hyperparameters that were optimal for the permutation invariant task.
In Table~\ref{tab:mnist_conv_results}, this model is referred to as
Conv-FC.

With the second network, which was inspired by
ConvPool-CNN-C from \citet{DBLP:journals/corr/SpringenbergDBR14}, we
only tested the $\Gamma$-model. The MNIST classification task can
typically be solved with a smaller number of parameters than CIFAR-10,
for which this topology was originally developed, so we
modified the network by removing layers and reducing the number of
parameters in the remaining layers. In addition, we observed that
adding a small fully connected layer with 10 neurons on top of the
global mean pooling layer improved the results in the semi-supervised
task. We did not tune other parameters than the noise level, which was
chosen from $\{0.3, 0.45, 0.6\}$ using the validation set.
The exact architecture of this network is detailed
in Table~\ref{table:conv-models} in Appendix~\ref{sec:convmodels}.
It is referred to as Conv-Small since it is a smaller version of
the network used forthe  CIFAR-10 dataset.
}{
We tested two convolutional networks for the general MNIST
classification task and focused on the 100-label case.
The first network was a straightforward extension of the
fully connected network tested in the permutation invariant case.  We
turned the first fully connected layer into a convolution with
26-by-26 filters, resulting in a 3-by-3 spatial map of 1000 features.
Each of the nine spatial locations was processed independently by a
network with the same structure as in the previous section, finally
resulting in a 3-by-3 spatial map of 10 features. These were pooled
with a global mean-pooling layer. We used the same
hyperparameters that were optimal for the permutation invariant task.
In Table~\ref{tab:mnist_conv_results}, this model is referred to as
Conv-FC.

With the second network, which was inspired by
ConvPool-CNN-C from \citet{DBLP:journals/corr/SpringenbergDBR14}, we
only tested the $\Gamma$-model.
The exact architecture of this network is detailed in
\s{Table~\ref{table:conv-models} in Appendix~\ref{sec:convmodels}}
{the supplementary material or Ref.~\cite{rasmus2015semi}}.
It is referred to as Conv-Small since it is a smaller version of
the network used for the CIFAR-10 dataset.
}

\begin{table}[t]
 \caption{CNN results for MNIST
 }
\begin{center}
 \begin{tabular}{llll}
   Test error without data augmentation \% with \# of used labels & 100 & all \\
   \hline
   EmbedCNN \citep{weston2012deep} & 7.75 &  \\
   SWWAE \citep{zhao2015stacked} & 9.17 & 0.71 \\
   \hline
   Baseline: Conv-Small, supervised only & 6.43 ($\pm$ 0.84)  & 0.36  \\
   Conv-FC & 0.99 ($\pm$ 0.15) & \\
   Conv-Small, $\Gamma$-model & {\bf 0.89} ($\pm$ 0.50) &  \\
\end{tabular}
 \end{center}
  % TODO: NIPS reviews: explain why some cells are empty
 \label{tab:mnist_conv_results}
\end{table}
% results/arxiv_mnist_conv_gamma_topctrue_100_60k6
% results/arxiv_mnist_conv_sup_ref_topctrue_100_60k3

The results in Table~\ref{tab:mnist_conv_results} confirm that even
the single convolution on the bottom level improves the results over
the fully connected network. More convolutions improve the
$\Gamma$-model significantly, although the \s{high variance of the results
suggests that the model still suffers from confirmation bias}
{variance is still high}. The
Ladder network with denoising targets on every level converges much
more reliably. Taken together, these results suggest that combining
the generalization ability of convolutional networks%
\footnote{In general, convolutional networks excel in the MNIST
classification task. The performance of the fully supervised Conv-Small
with all labels is in line with the literature and is provided as a rough
reference only (only one run, no attempts to optimize, not available
in the code package).}
and efficient unsupervised learning of the full Ladder network would
have resulted in even better performance but this was left for future
work.

\subsection{Convolutional networks on CIFAR-10}

\s{
The CIFAR-10 dataset consists of small 32-by-32 RGB images from 10
classes. There are 50,000 labeled samples for training and 10,000 for
testing. Like the MNIST dataset, it has been used for testing
semi-supervised learning so we decided to test the simple
$\Gamma$-model with a convolutional network that has been reported to
perform well in the standard supervised setting with all labels.  We
tested a few model architectures and selected ConvPool-CNN-C by
\citet{DBLP:journals/corr/SpringenbergDBR14}. We also evaluated the
strided convolutional version by
\citet{DBLP:journals/corr/SpringenbergDBR14}, and while it performed
well with all labels, we found that the max-pooling version overfitted
less with fewer labels, and thus used it.

The main differences to ConvPool-CNN-C are the use of Gaussian noise
instead of dropout and the convolutional per-channel batch
normalization following \citet{Ioffe15icml}. While dropout was useful
with all labels, it did not seem to offer any advantage over additive
Gaussian noise with fewer labels.
For a more detailed description of the 
model, please refer to model Conv-Large in Table~\ref{table:conv-models}.

While testing the purely supervised model performance with a limited number of labeled
samples ($N=4000$), we found out that the model overfitted quite
severely: the training error for most samples decreased so much that the
network effectively learned nothing from them as the network was
already very confident about their classification. The network was
equally confident about validation samples even when they were
misclassified. We noticed that we could regularize the network by
stripping away the scaling parameter $\boldsymbol\beta^{(L)}$ from the
last layer. This means that the variance of the input to the softmax
is restricted to unity. We also used this setting with the
corresponding $\Gamma$-model although the denoising target already
regularizes the network significantly and the improvement was not as
pronounced.
}{
The CIFAR-10 dataset consists of small 32-by-32 RGB images from 10
classes. There are 50,000 labeled samples for training and 10,000 for
testing. We decided to test the simple
$\Gamma$-model with the convolutional architecture ConvPool-CNN-C by
\citet{DBLP:journals/corr/SpringenbergDBR14}. The main differences to
ConvPool-CNN-C are the use of Gaussian noise
instead of dropout and the convolutional per-channel batch
normalization following \citet{Ioffe15icml}.
For a more detailed description of the 
model, please refer to model Conv-Large in \s{Table~\ref{table:conv-models}}{the supplementary material}.

% In order to alleviate overfitting with the purely supervised model,
% we stripped away the batch normalization scaling
% parameter $\boldsymbol\beta^{(L)}$ from the
% last layer. Based on the validation runs, we found that this improved the results.
}

The hyperparameters (noise level, denoising cost multipliers, and number
of epochs) for all models were optimized using $M=40,000$ samples
for training and the remaining $10,000$ samples for validation.
After the best hyperparameters were selected, the final model was trained
with these settings on all the $M=50,000$ samples. All experiments were run with with four different random initializations of the weight matrices and data splits.
We applied global contrast normalization and whitening
following \citet{Goodfellow_maxout_2013}, but no data augmentation was used.

\begin{table}[t]
 \caption{Test results for CNN on CIFAR-10 dataset without data augmentation}
\begin{center}
 \begin{tabular}{llll}
   Test error \% with \# of used labels & 4~000 & All \\
   \hline
   All-Convolutional ConvPool-CNN-C \citep{DBLP:journals/corr/SpringenbergDBR14} &   & 9.31 \\
   Spike-and-Slab Sparse Coding \citep{goodfellow2012large} & 31.9  & \\
   \hline
   Baseline: Conv-Large, supervised only & 23.33 ($\pm$ 0.61) & 9.27 \\
   Conv-Large, $\Gamma$-model & {\bf 20.40} ($\pm$ 0.47) & \\
\end{tabular}
 \end{center}
 \label{tab:cifar10_results}
\end{table}
 % TODO: NIPS reviews: explain why some cells are empty

The results are shown in Table~\ref{tab:cifar10_results}. The supervised
reference was obtained with a model closer to the original ConvPool-CNN-C
in the sense that dropout rather than additive Gaussian noise was used
for regularization.%
\footnote{Same caveats hold for this fully supervised reference result
  for all labels as with MNIST: only one run, no attempts to optimize,
  not available in the code package.}
We spent some time tuning the regularization of our fully supervised
baseline model for $N=4000$ labels and indeed, its results exceed the
previous state of the art. This tuning was important to make sure that
the improvement offered by the denoising target of the $\Gamma$-model
is not a sign of a poorly regularized baseline model. Although the
improvement is not as dramatic as with the MNIST experiments, it came with
a very simple addition to standard supervised training.

% % % % % % % % % % % % % % % % % % % % % % % % % % % % 
\section{Related Work}
% % % % % % % % % % % % % % % % % % % % % % % % % % % % 

\s{
Early works on semi-supervised learning \citep{mclachlan1975,titterington1985} 
proposed an approach where inputs $\vx$ are first assigned to clusters, and 
each cluster has its class label. Unlabeled data would affect the shapes and 
sizes of the clusters, and thus alter the classification result. This approach 
can be reinterpreted as input vectors being corrupted copies $\tilde{\vx}$ of 
the ideal input vectors $\vx$ (the cluster centers), and the classification 
mapping being split into two parts: first denoising $\tilde{\vx}$ into $\vx$ 
(possibly probabilistically), and then labeling $\vx$.

It is well known \citep[see, e.g.,][]{zhang2000} that when a probabilistic
model that directly estimates $P(y\mid \vx)$ is being trained, unlabeled data cannot help. One 
way to study this is to assign probabilistic labels $q(y(n))=P(y(n)\mid \vx(n))$ 
to unlabeled inputs $\vx(n)$ and try to train $P(y\mid \vx)$ using those 
labels: it can be shown \citep[see, e.g.,][Eq. (31)]{Raiko2015techniques} that the 
gradient will vanish. There are different ways of circumventing this 
phenomenon by adjusting the assigned labels $q(y(n))$. These are all related to the $\Gamma$-model.

% TODO: Fix y_t vs. y(n)
Label propagation methods \citep{szummer2002partially} estimate $P(y\mid \vx)$, 
but adjust probabilistic labels $q(y(n))$ on the basis of the assumption that the
nearest neighbors are likely to have the same label. The labels start to 
propagate through regions with high-density $P(\vx)$. 
The $\Gamma$-model implicitly assumes
that the labels are uniform in the vicinity of a clean input
since corrupted inputs need to produce the same label. This produces a similar 
effect: the labels start to propagate through regions with high density 
$P(\vx)$. \citet{weston2012deep} explored deep versions of label propagation.

Co-training \citep{blum1998combining} assumes we have multiple views on $\vx$,
say $\vx=(\vx^{(1)},\vx^{(2)})$. When we train classifiers for the 
different views, we know that even for the unlabeled data, the true label 
is the same for each view. Each view produces its own probabilistic labeling 
$q^{(j)}(y(n))=P(y(n)\mid \vx(n)^{(j)})$ and their combination $q(y(n))$ can be 
fed to train the individual classifiers. If we interpret having several 
corrupted copies of an input as different views on it, we see the relationship 
to the proposed method.

\citet{lee2013pseudo} adjusts the assigned labels $q(y(n))$ by rounding the
probability of the most likely class to one and others to zero. The training
starts by trusting only the true labels and then gradually increasing the
weight of the so-called {\it pseudo-labels}. Similar scheduling could be
tested with our $\Gamma$-model as it seems to suffer from confirmation bias.
It may well be that the denoising cost which is optimal at the beginning
of the learning is smaller than the optimal one at later stages of learning.

\citet{dosovitskiy2014discriminative} pre-train a convolutional network with
unlabeled data by treating each clean image as its own class. During training,
the image is corrupted by transforming its location, scaling, rotation, contrast,
and color. This helps to find features that are invariant to the
transformations that are used. Discarding the last classification layer and replacing it with
a new classifier trained on real labeled data leads to surprisingly good
experimental results.

There is an interesting connection between our $\Gamma$-model and the
contractive cost used by \citet{rifai2011manifold}: a linear denoising
function $\hat z^{(L)}_i = a_i \tilde{z}^{(L)}_i + b_i$, where $a_i$
and $b_i$ are parameters, turns the denoising cost into a stochastic
estimate of the contractive cost.

Recently \citet{miyato2015} achieved impressive results with a
regularization method that is similar to the idea of contractive cost.
They required the output of the network to change as little as
possible close to the input samples. As this requires no labels, they
were able to use unlabeled samples for regularization. While their
semi-supervised results were not as good as ours with a denoising
target on the input layer, their results with full labels come very
close. Their cost function is on the last layer which suggests that
the approaches are complementary and could be combined, potentially
improving the results further.

So far we have reviewed semi-supervised methods which have an
unsupervised cost function on the output layer only and therefore are
related to our $\Gamma$-model.
We will now move to other semi-supervised methods that concentrate on
modeling the joint distribution of the inputs and the labels.

The Multi-prediction deep Boltzmann machine (MP-DBM) \citep{goodfellow2013multi} is 
a way to train a DBM with backpropagation through variational inference. The 
targets of the inference include both supervised targets (classification) and 
unsupervised targets (reconstruction of missing inputs) that are used in 
training simultaneously. The connections through the inference network are 
somewhat analogous to our lateral connections. Specifically, there are 
inference paths from observed inputs to reconstructed inputs that do not go all 
the way up to the highest layers. Compared to our approach, MP-DBM requires an 
iterative inference with some initialization for the hidden activations, 
whereas in our case, the inference is a simple single-pass feedforward 
procedure.

The Deep AutoRegressive Network \citep{gregor2014deep} is an unsupervised 
method for learning representations that also uses lateral connections in the 
hidden representations. The connectivity within the layer is rather different 
from ours, though: each unit $h_i$ receives input from the preceding units 
$h_1\dots h_{i-1}$, whereas in our case each unit $\hat{z}_i$ receives input 
only from $z_i$. Their learning algorithm is based on approximating a gradient 
of a description length measure, whereas we use a gradient of a simple loss 
function.

\citet{kingma2014semi} proposed deep generative models for semi-supervised 
learning, based on variational autoencoders. Their models can be trained 
with the variational EM algorithm, stochastic gradient variational Bayes, or 
stochastic backpropagation. They also experimented on a stacked version 
(called M1+M2) where the bottom autoencoder M1 reconstructs the input data, and 
the top autoencoder M2 can concentrate on classification and on reconstructing 
only the hidden representation of M1. The stacked version performed the best, 
hinting that it might be important not to carry all the information up to the 
highest layers. Compared with the Ladder network, an interesting point is
that the variational autoencoder computes the posterior estimate of
the latent variables with the encoder alone while the Ladder network uses
the decoder too to compute an implicit posterior approximate (the encoder
provides the likelihood part, which gets combined with the prior). It
will be interesting to see whether the approaches can be combined. A
Ladder-style decoder might provide the posterior and another decoder
could then act as the generative model of variational autoencoders.

\citet{zeiler2011adaptive} train deep convolutional autoencoders in a 
manner comparable to ours. They define max-pooling operations in the 
encoder to feed the max function upwards to the next layer, while the 
argmax function is fed laterally to the decoder. The network is trained 
one layer at a time using a cost function that includes a pixel-level 
reconstruction error, and a regularization term to promote sparsity.
\citet{zhao2015stacked} use a similar structure and call it the stacked 
what-where autoencoder (SWWAE).
Their network is trained simultaneously to minimize a combination of the 
supervised cost and reconstruction errors on each level, just like ours.

Recently \citet{bengio2014auto} proposed target propagation as an
alternative to backpropagation. The idea is to base learning not on
errors and gradients but on expectations. This is very similar to the
idea of denoising source separation and therefore resembles the
propagation of expectations in the decoder of the Ladder network.  In
the Ladder network, the additional lateral connections between the
encoder and the decoder play an important role and it remains to
be seen whether the lateral connections are compatible with target
propagation. Nevertheless, it is an interesting possibility that while
the Ladder network includes two mechanisms for propagating
information, backpropagation of gradients and forward propagation of
expectations in the decoder, it may be possible to rely solely on the
latter, thus avoiding problems related to the propagation of gradients through
many layers, such as exploding gradients.
% END ARXIV
}{
% BEGIN NIPS
Early works on semi-supervised learning \citep{mclachlan1975,titterington1985} 
proposed an approach where inputs $\vx$ are first assigned to clusters, and 
each cluster has its class label. Unlabeled data would affect the shapes and 
sizes of the clusters, and thus alter the classification result.
% This approach 
% can be reinterpreted as input vectors being corrupted copies $\tilde{\vx}$ of 
% the ideal input vectors $\vx$ (the cluster centers), and the classification 
% mapping being split into two parts: first denoising $\tilde{\vx}$ into $\vx$ 
% (possibly probabilistically), and then labeling $\vx$.
%
% It is well known \citep[see, e.g.,][]{zhang2000} that when training a probabilistic
% model that directly estimates $P(y\mid \vx)$, unlabeled data cannot help.
% However, one can assign probabilistic labels $q(y_t)=P(y_t\mid \vx_t)$ 
% to unlabeled inputs $\vx_t$ and adjust the assigned labels $q(y_t)$ to use
% as labels for the unlabeled data.
% These are all related to the $\Gamma$-model.
%
Label propagation methods \citep{szummer2002partially} estimate $P(y\mid \vx)$, 
but adjust probabilistic labels $q(y(n))$ on the basis of the assumption that the
nearest neighbors are likely to have the same label.
% The labels start to 
% propagate through regions with high density $P(\vx)$. 
% The $\Gamma$-model implicitly assumes
% that the labels are uniform in the vicinity of a clean input
% since corrupted inputs need to produce the same label. This produces a similar 
% effect: The labels start to propagate through regions with high density 
% $P(\vx)$.
\citet{weston2012deep} explored deep versions of label propagation.

% Co-training \citep{blum1998combining} assumes we have multiple views on $\vx$,
% say $\vx=(\vx^{(1)},\vx^{(2)})$.
% When we train classifiers for the 
% different views, we know that even for the unlabeled data, the true label 
% is the same for each view.
% Each view produces its own probabilistic labeling 
% $q^{(j)}(y_t)=P(y_t\mid \vx_t^{(j)})$ and their combination $q(y_t)$ can be 
% fed to train the individual classifiers. If we interpret having several 
% corrupted copies of an input as different views on it, we see the relationship 
% to the proposed method.

% \citet{lee2013pseudo} adjusts the assigned labels $q(y_t)$ by rounding the
% probability of the most likely class to one and others to zero. The training
% starts by trusting only the true labels and then gradually increasing the
% weight of the so called {\it pseudo-labels}.
% Similar scheduling could be
% tested with our $\Gamma$-model as it seems to suffer from confirmation bias.
% It may well be that the denoising cost which is optimal in the beginning
% of the learning is smaller than the optimal at later stages of learning.

% \citet{dosovitskiy2014discriminative} pre-train a convolutional network with
% unlabeled data by treating each clean image as its own class. During training,
% the image is corrupted by transforming its location, scaling, rotation, contrast,
% and color. This helps to find features that are invariant to the used
% transformations. Discarding the last classification layer and replacing it with
% a new classifier trained on real labeled data leads to surprisingly good
% experimental results.

There is an interesting connection between our $\Gamma$-model and the
contractive cost used by \citet{rifai2011manifold}: a linear denoising
function $\hat z^{(L)}_i = a_i \tilde{z}^{(L)}_i + b_i$, where $a_i$
and $b_i$ are parameters, turns the denoising cost into a stochastic
estimate of the contractive cost. In other words, our $\Gamma$-model
seems to combine clustering and label propagation with regularization
by contractive cost.

Recently \citet{miyato2015} achieved impressive results with a
regularization method that is similar to the idea of contractive cost.
They required the output of the network to change as little as
possible close to the input samples. As this requires no labels, they
were able to use unlabeled samples for regularization.
% While their
% semi-supervised results were not as good as ours with a denoising
% target at the input layer, their results with full labels come very
% close. Their cost function is at the last layer which suggests that
% the approaches are complementary and could be combined, potentially
% further improving the results.

% So far we have reviewed semi-supervised methods which have an
% unsupervised cost function at the output layer only and therefore are
% related to our $\Gamma$-model.
% We will now move to other semi-supervised methods that concentrate on
% modeling the joint distribution of the inputs and the labels.

The Multi-prediction deep Boltzmann machine (MP-DBM) \citep{goodfellow2013multi} is 
a way to train a DBM with backpropagation through variational inference. The 
targets of the inference include both supervised targets (classification) and 
unsupervised targets (reconstruction of missing inputs) that are used in 
training simultaneously. The connections through the inference network are 
somewhat analogous to our lateral connections. Specifically, there are 
inference paths from observed inputs to reconstructed inputs that do not go all 
the way up to the highest layers. Compared to our approach, MP-DBM requires an 
iterative inference with some initialization for the hidden activations, 
whereas in our case, the inference is a simple single-pass feedforward 
procedure.

% The Deep AutoRegressive Network \citep{gregor2014deep} is an unsupervised 
% method for learning representations that also uses lateral connections in the 
% hidden representations. The connectivity within the layer is rather different 
% from ours, though: Each unit $h_i$ receives input from the preceding units 
% $h_1\dots h_{i-1}$, whereas in our case each unit $\hat{z}_i$ receives input 
% only from $z_i$. Their learning algorithm is based on approximating a gradient 
% of a description length measure, whereas we use a gradient of a simple loss 
% function.

\citet{kingma2014semi} proposed deep generative models for semi-supervised 
learning, based on variational autoencoders.
Their models can be trained 
with the variational EM algorithm, stochastic gradient variational Bayes, or 
stochastic backpropagation.
% They also experimented on a stacked version 
% (called M1+M2) where the bottom autoencoder M1 reconstructs the input data, and 
% the top autoencoder M2 can concentrate on classification and on reconstructing 
% only the hidden representation of M1. The stacked version performed the best, 
% hinting that it might be important not to carry all the information up to the 
% highest layers.
Compared with the Ladder network, an interesting point is
that the variational autoencoder computes the posterior estimate of
the latent variables with the encoder alone while the Ladder network uses
the decoder too to compute an implicit posterior approximate (the encoder
provides the likelihood part, which gets combined with the prior).
% It
% will be interesting to see whether the approaches can be combined. A
% Ladder-style decoder might provide the posterior and another decoder
% could then act as the generative model of variational autoencoders.

\citet{zeiler2011adaptive} train deep convolutional autoencoders in a 
manner comparable to ours. They define max-pooling operations in the 
encoder to feed the max function upwards to the next layer, while the 
argmax function is fed laterally to the decoder. The network is trained 
one layer at a time using a cost function that includes a pixel-level 
reconstruction error, and a regularization term to promote sparsity.
\citet{zhao2015stacked} use a similar structure and call it the stacked 
what-where autoencoder (SWWAE).
Their network is trained simultaneously to minimize a combination of the 
supervised cost and reconstruction errors on each level, just like ours.

% Recently \citet{bengio2014auto} proposed target propagation as an
% alternative to backpropagation. The idea is to base learning not on
% errors and gradients but on expectations. This is very similar to the
% idea of denoising source separation and therefore resembles the
% propagation of expectations in the decoder of the Ladder network.  In
% the Ladder network, the additional lateral connections between the
% encoder and the decoder play an important role and it will remain to
% be seen whether the lateral connections are compatible with target
% propagation. Nevertheless, it is an interesting possibility that while
% the Ladder network includes two mechanisms for propagating
% information, backpropagation of gradients and forward propagation of
% expectations in the decoder, it may be possible to rely solely on the
% latter, avoiding problems related to propagation of gradients through
% many layers, such as exploding gradients.
}

% % % % % % % % % % % % % % 
\section{Discussion}
% % % % % % % % % % % % % % 

We showed how a simultaneous unsupervised learning task improves CNN and MLP networks reaching
the state of the art in various semi-supervised learning tasks. In particular, the performance
obtained with very small numbers of labels is much better than previous published results,
which shows that the method is capable of making good use of unsupervised learning. However,
the same model also achieves state-of-the-art results and a
significant improvement over the baseline model with full labels in
permutation invariant MNIST classification, which suggests that the
unsupervised task does not disturb supervised learning.

The proposed model is simple and easy to implement with many existing feedforward
architectures, as the training is based on backpropagation from a simple cost 
function. It is quick to train and the convergence is fast, thanks to
batch normalization.

Not surprisingly, the largest improvements in performance were observed in
models which have a large number of parameters relative to the number
of available labeled samples. With CIFAR-10, we started with a model
which was originally developed for a fully supervised task. This has
the benefit of building on existing experience but it may well be that
the best results will be obtained with models which have far more
parameters than fully supervised approaches could handle.

An obvious future line of research will therefore be to study what kind of
encoders and decoders are best suited to the Ladder network. In this
work, we made very small modifications to the encoders, whose
structure has been optimized for supervised learning, and we designed
the parametrization of the vertical mappings of the decoder to mirror
the encoder: the flow of information is just reversed. There is
nothing preventing the decoder from having a different structure than the
encoder.
%Also, there were lateral connections from the encoder to the
%decoder on every layer and on every pooling operation. The miniature
%MLPs used for every denoising function g gives the decoder enough
%capacity to invert the mappings but the same effect could have been
%accomplished by not requiring the decoder to match the activations of
%the encoder on every layer.

An interesting future line of research will be the
extension of the Ladder networks to the temporal domain. While
datasets with millions of labeled samples for still images exist, it
is prohibitively costly to label thousands of hours of video streams.
The Ladder networks can be scaled up easily and therefore offer an
attractive approach for semi-supervised learning in such large-scale
problems.

\section*{Acknowledgements}

We have received comments and help from a number of colleagues who
all deserve to be mentioned but we wish to thank especially Yann
LeCun, Diederik Kingma, Aaron Courville, Ian Goodfellow, S{\o}ren S{\o}nderby,
Jim Fan, and Hugo Larochelle for their
helpful comments and suggestions. The software for the simulations for
this paper was based on
Theano \s{\citep{bastien-theano,bergstra-theano-scipy}}{\citep{bastien-theano}} and
Blocks \citep{blocks_MerrienboerBDSW15}.
We also acknowledge the computational resources provided by the Aalto
Science-IT project.
The Academy of Finland has supported Tapani Raiko.
\s{}{\small}{
\bibliography{ladder}
\s{\bibliographystyle{natbib}}{\bibliographystyle{unsrtnat}}
% NIPS: unsrtnat
}
\s{}{\normalsize}

%%%%%%%%%%%%%%%%%%%%%%%%%%%%%%%%%
\newpage
\appendix
%%%%%%%%%%%%%%%%%%%%%%%%%%%%%%%%%

\s{}{
\section{Details of the Implementation of the Model}
The implementation details of the decoder for the MLP and convolutional models are
presented in the following section.

% % % % % % % % % % % % % % 
\subsection{Decoder for Unsupervised Learning}
% % % % % % % % % % % % % % 
\label{sec:model.unsuper}

When designing a suitable decoder to support unsupervised learning, we had to
make a choice as to what kind of distributions of the latent variables the decoder would optimally
be able to denoise. Based on preliminary analyses,
we ultimately ended up choosing a parametrization that
supports optimal denoising of latent variables that are independently distributed Gaussian
variables conditional on the values of the latent variables of the layer above. The distribution of the latent variables
$\z l$ is therefore assumed to follow the distribution
$$p(\z l \mid \z {l+1}) = \prod_i p(z_i^{(l)} \mid \z {l+1})$$
where $p(z_i^{(l)} \mid \z {l+1})$
are conditionally independent Gaussian distributions of latent variables $z_i$ in layer $l$.

It can
be shown that the functional form of $z^{(l)}_i = g_i(\tilde{z}^{(l)}_i, u_i^{(l)})$ has to be linear
in order to minimize the denoising cost with the assumption that both the noise and the
latent variable have a Gaussian distribution \citep[][Section 4.1]{valpola2015ladder}.
The denoising function was therefore chosen to have the form
\begin{equation}
\hat z^{(l)}_i = g_i(\tilde{z}^{(l)}_i, u_i^{(l)}) = \left(\tilde{z}^{(l)}_i - \mu_i (u_i^{(l)})\right) \upsilon_i (u_i^{(l)}) + \mu_i (u_i^{(l)}) \,
\label{eq:g_func}
\end{equation}
where $u_i^{(l)}$ is a function of $\hz{l+i}$.
Somewhat arbitrarily, we choose the functional form pf $\u l$ to be
a vertical mapping from $\hz {l+1}$, which is then batch normalized:
$$
	\u l = \mathtt{batchnorm}(\V {l+1} \hz {l+1}) \, ,
$$
where the matrix $\V l$ has the same dimension as the transpose of $\W l$ on the
encoder side. The projection vector $\u l$ therefore has the same dimensionality as $\z l$.
Furthermore, the functions $\mu_i (u_i^{(l)})$ and $\upsilon_i (u_i^{(l)})$ are modeled
as expressive nonlinearities:
$$
\mu_i (u_i^{(l)}) = a_{1,i}^{(l)} \mathtt{sigmoid}(a_{2,i}^{(l)} u_i^{(l)} + a_{3,i}^{(l)}) + a_{4,i}^{(l)} u_i^{(l)} + a_{5,i}^{(l)}
$$
$$
\upsilon_i (u_i^{(l)}) = a_{6,i}^{(l)} \mathtt{sigmoid}(a_{7,i}^{(l)} u_i^{(l)} + a_{8,i}^{(l)}) + a_{9,i}^{(l)} u_i^{(l)} + a_{10,i}^{(l)},
$$
where $a_{1,i}^{(l)} \dots a_{10,i}^{(l)}$ are the trainable parameters of the nonlinearity
for each neuron $i$ in each layer $l$.

For the lowest layer, $\hat{\x} = \hz 0$ and $\tilde \x = \tz 0$ by definition,
and for the highest layer we chose $\u L = \tilde \y$. This allows the highest-layer
denoising function to utilize prior information about the classes being mutually
exclusive which seems to improve convergence in cases where there are very few
labeled samples.

One interpretation of this formulation is that we are modeling the distribution of $\z l$ as a mixture of
Gaussians with diagonal covariance matrices, where the value of the above layer
$\z {l+1}$ modulates the form of the Gaussian that $\z l$ is distributed as.
In this parametrization all correlations, nonlinearities
and non-Gaussianities in the latent variables $\z l$ have to be represented by modulations
from above layers for optimal denoising.
The parametrizaiton therefore encourages the decoder to find representations $\z l$ that have
high mutual information with $\z {l+1}$. This is crucial as it allows
supervised learning to have an indirect influence on the representations
learned by the unsupervised decoder: any abstractions selected by supervised
learning will bias the lower levels to find more representations which
carry information about the same abstractions.

\citet{Rasmus15arxiv} showed that modulated connections in $g$ are
crucial for allowing the decoder to recover discarded details from the
encoder and thus for allowing invariant representations to develop.
The proposed parametrization can represent such modulation but also
traditional top-down decoder connections that are normally used in dAEs.
We also tested alternative formulations for the denoising function,
the results of which can be found in Appendix~\ref{sec:comparison}.

The cost function for the unsupervised path is the mean squared reconstruction
error per neuron, but there is a slight twist which we found to be important.
Batch normalization has useful properties,
but it also introduces noise which affects both the clean and corrupted
encoder pass. This noise is highly correlated between $\z l$ and
$\tz l$ because the noise derives from the statistics of the samples that
happen to be in the same minibatch. This highly correlated noise in $\z l$
and $\tz l$ biases the denoising functions to be simple copies%
\footnote{The whole point of using \textit{denoising} autoencoders rather
than regular autoencoders is to prevent skip connections from short-circuiting
the decoder and force the decoder to learn meaningful abstractions which
help in denoising.} $\hz l \approx \tz l$.

The solution we found was to implicitly use the projections $\z l_{\mathrm{pre}}$
as the target for denoising and scale the cost function in such a way that
the term appearing in the error term is the batch normalized $\z l$ instead.
For the moment, let us see how that works for a scalar case:
\begin{eqnarray*}
	\frac{1}{\sigma^2} \left\Vert z_{\mathrm{pre}} - \hat z \right\Vert ^2 & = &
    \left\Vert \frac{z_{\mathrm{pre}} - \mu}{\sigma} - \frac{\hat{z} - \mu}{\sigma}\right\Vert^2 =
    \left\Vert z - \hat{z}_{\mathrm{BN}}\right\Vert^2 \\
    z & = & \mathtt{batchnorm}(z_{\mathrm{pre}}) = \frac{z_{\mathrm{pre}} - \mu}{\sigma} \\
    \hat{z}_{\mathrm{BN}} & = & \frac{\hat{z} - \mu}{\sigma} \, ,
\end{eqnarray*}

where $\mu$ and $\sigma$ are the batch mean and batch std of $z_{\mathrm{pre}}$,
respectively, that were used in batch normalizing $z_{\mathrm{pre}}$ into $z$. 
The unsupervised denoising cost function $C_\mathrm{d}$ is thus
\begin{equation}
C_\mathrm{d} = \sum_{l=0}^L \lambda_l C_\mathrm{d}^{(l)}
= \sum_{l=0}^L \frac {\lambda_l} {Nm_l} \sum_{n=1}^N \left\Vert \z l(n) - \hz l_{\mathrm{BN}}(n) \right\Vert^2,
\label{eq:cost}
\end{equation}
where $m_l$ is the layer's width, N the number of training samples, and the hyperparameter
$\lambda_l$ a layer-wise multiplier determining the importance of the denoising cost.

The model parameters $\W l,\boldsymbol{\gamma}^{(l)},\boldsymbol{\beta}^{(l)},\V l,\mathbf{a}_i^{(l)}$
can be trained simply by using the backpropagation algorithm to optimize
the total cost $C = C_\mathrm{c} + C_\mathrm{d}$. The feedforward pass of the full Ladder
network is listed in Algorithm~\ref{alg:ff}. Classification results are
read from the $\y$ in the clean feedforward path.

\subsection{Variations}
\label{sec:variations_appendix}

Section \ref{sec:model.unsuper} detailed how to build a decoder for
the Ladder network to match a fully connected encoder.
It is easy to extend the same approach to other
encoders, for instance, convolutional neural networks (CNN). For the
decoder of fully connected networks we used vertical mappings whose
shape is a transpose of the encoder mapping. The same treatment works
for the convolution operations: in the networks we have tested in this
paper, the decoder has convolutions whose parametrization mirrors the
encoder and effectively just reverses the flow of information. As the
idea of convolution is to reduce the number of parameters by weight
sharing, we applied this to the parameters of the denoising function
g, too.

Many convolutional networks use pooling operations with stride, that
is, they downsample the spatial feature maps. The decoder needs to
compensate this with a corresponding upsampling. There are several
alternative ways to implement this and in this paper we chose the
following options: 1) on the encoder side, pooling operations are
treated as separate layers with their own batch normalization and
linear activations function and 2) the downsampling of the pooling on
the encoder side is compensated by upsampling with copying on the
decoder side. This provides multiple targets for the decoder to match,
helping the decoder to recover the information lost on the encoder
side.

It is worth noting that a simple special case of the decoder is a model where
$\lambda_l = 0$ when $l < L$. This corresponds to a denoising cost only on the top layer
and means that most of the decoder can be omitted. This model, which we call the $\Gamma$-model
due to the shape of the graph, is useful as it can easily be plugged into any feedforward
network without decoder implementation. In addition, the $\Gamma$-model is the same
for MLPs and convolutional neural networks. The encoder in the $\Gamma$-model
still includes both the clean and the corrupted paths as in the full ladder.
}

% % % % % % % % % % % % % % 
\section{Specification of the convolutional models}
% % % % % % % % % % % % % % 
\label{sec:convmodels}

%convv:96:3:1:1-convf:96:3:1:1-convf:96:3:1:1-maxpool:2:2-convv:192:3:1:1-convf:192:3:1:1-convv:192:3:1:1-maxpool:2:2-convv:192:3:1:1-convv:192:1:1:1-convv:10:1:1:1-meanpool:6:6
% convf:32:5:1:1-maxpool:2:2-conv:64:3:1:1-convf:64:3:1:1-maxpool:2:2-conv:128:3:1:1-convv:10:1:1:1-meanpool:6:6-fc:10

\begin{table}[h]
\caption{ConvPool-CNN-C by \citet{DBLP:journals/corr/SpringenbergDBR14} and our networks based on it.}
\label{table:conv-models}
\begin{center}
\begin{small}
\begin{tabular}{l|l|l}
\multicolumn{3}{c}{\bf Model} \\
\hline
ConvPool-CNN-C         &  Conv-Large (for CIFAR-10) & Conv-Small (for MNIST) \\
\hline
\multicolumn{3}{c}{Input $32 \times 32$ or $28 \times 28$ RGB or monochrome image} \\
\hline
$3 \times 3$ conv. $96$ ReLU  & $3 \times 3$ conv. $96$ BN LeakyReLU  & $5 \times 5$ conv. $32$ ReLU  \\
$3 \times 3$ conv. $96$ ReLU  & $3 \times 3$ conv. $96$ BN LeakyReLU  &   \\
$3 \times 3$ conv. $96$ ReLU  & $3 \times 3$ conv. $96$ BN LeakyReLU  &   \\
\hline
$3 \times 3$ max-pooling stride $2$  & $2 \times 2$ max-pooling stride $2$ BN & $2 \times 2$ max-pooling stride $2$ BN  \\
\hline
$3 \times 3$ conv. $192$ ReLU & $3 \times 3$ conv. $192$ BN LeakyReLU & $3 \times 3$ conv. $64$ BN ReLU \\
$3 \times 3$ conv. $192$ ReLU & $3 \times 3$ conv. $192$ BN LeakyReLU & $3 \times 3$ conv. $64$ BN ReLU \\
$3 \times 3$ conv. $192$ ReLU & $3 \times 3$ conv. $192$ BN LeakyReLU &  \\
\hline
$3 \times 3$ max-pooling stride $2$  & $2 \times 2$ max-pooling stride $2$ BN & $2 \times 2$ max-pooling stride $2$ BN  \\
\hline
$3 \times 3$ conv. $192$ ReLU & $3 \times 3$ conv. $192$ BN LeakyReLU & $3 \times 3$ conv. $128$ BN ReLU \\
$1 \times 1$ conv. $192$ ReLU & $1 \times 1$ conv. $192$ BN LeakyReLU & \\
$1 \times 1$ conv. $10$ ReLU & $1 \times 1$ conv. $10$ BN LeakyReLU &  $1 \times 1$ conv. $10$ BN ReLU \\
\hline
global meanpool & global meanpool BN & global meanpool BN \\
\hline
 &  &  fully connected $10$ BN \\
\hline
\multicolumn{3}{c}{10-way softmax} \\
\end{tabular}
\end{small}
\end{center}
\end{table}

Here we describe two model structures, Conv-Small and Conv-Large, that
were used for the MNIST and CIFAR-10 datasets, respectively.
They were both inspired by ConvPool-CNN-C by
\citet{DBLP:journals/corr/SpringenbergDBR14}.
Table~\ref{table:conv-models} details the model architectures and differences between the models in this work and ConvPool-CNN-C. It is noteworthy that this 
architecture does not use any fully connected layers, but replaces them 
with a global mean pooling layer just before the softmax function. The main differences between our models and ConvPool-CNN-C are the use of Gaussian noise instead 
of dropout and the convolutional per-channel batch normalization following 
\citet{Ioffe15icml}. We also used 2x2 stride 2 max-pooling instead of 3x3 
stride 2 max-pooling. LeakyReLU was used to speed up training, as mentioned
by \citet{DBLP:journals/corr/SpringenbergDBR14}. We utilized batch
normalization to all layers, including the pooling layers. Gaussian noise
was also added to all layers, instead of applying dropout in only some
of the layers as with ConvPool-CNN-C.

% % % % % % % % % % % % % % 
\section{Formulation of the Denoising Function}
% % % % % % % % % % % % % % 
\label{sec:comparison}

The denoising function $g$ tries to map the clean $\z l$ to the reconstructed $\hz l$,
where $\hz l = g(\tz l, \hz {l+1})$. The reconstruction is therefore
based on the corrupted value and the reconstruction of the layer above.

An optimal functional form of $g$ depends on the conditional distribution $p(\z l \mid \z {l+1})$
that we want the model to be able to denoise.
For example, if the distribution $p(\z l \mid \z {l+1})$ is Gaussian, the optimal function $g$, that is,
the function that achieves the lowest reconstruction error, 
is going to be linear with respect to $\tz l$ \citep[][Section 4.1]{valpola2015ladder}. This is
the parametrization that we chose on the basis of preliminary comparisons of different denoising function
parametrizations.

% When analyzing the distribution $p(\z l \mid \z {l+1})$ learned by a purely supervised
% network, it would therefore be desirable to parametrize $g$ in such a way as
% to be able to optimally denoise the kinds of distributions the network has found
% for the hidden activations.

% In our preliminary analyses of the distributions learned by the hidden layers,
% we found many different very non-Gaussian distributions that we wanted the $g$-function
% to be able to denoise. One example were bimodal distributions, that were often observed
% in the layer below the final classification layer. We could also observe that in many cases, the value
% of $\z {l+1}$ had an impact on $p(\z l \mid \z {l+1})$ beyond shifting the mean
% of the distribution, which led us to propose a form where the vertical connections from 
% $\hz {l+1}$ could modulate the horizontal connections from $\tz l$ instead of
% only additively shifting the distribution defined by $g$. This corresponds
% to letting the variance and other higher-order cumulants of $\z l$ depend on $\hz {l+1}$.

The proposed parametrization of the denoising function was therefore:
\begin{equation}
g(\tilde{z}, u) = \left(\tilde{z} - \mu (u)\right) \upsilon(u) + \mu(u) \, .
\end{equation}
We modeled both $\mu(u)$ and $\upsilon(u)$ with an expressive nonlinearity
\footnote{The parametrization can also be interpreted as a miniature MLP network}:
$\mu(u)= a_1 \mathtt{sigmoid}(a_2 u + a_3) + a_4 u + a_5$ and 
$\upsilon(u)= a_6 \mathtt{sigmoid}(a_7 u + a_8) + a_9 u + a_{10}$.
We have left out the superscript
$(l)$ and subscript $i$ in order not to clutter the equations.
Given $u$, this parametrization 
is linear with respect to $\tilde{z}$, and
both the slope and the bias depended nonlinearly on $u$.

In order to test whether the elements of the proposed function $g$
were necessary, we systematically removed components from $g$ or
replaced $g$ altogether and compared the resulting performance to the results obtained with the
original parametrization. We tuned the hyperparameters of each
comparison model separately using a grid search over some of the
relevant hyperparameters. However, the standard deviation of additive Gaussian
corruption noise was set to 0.3. This means that
the comparison does not include the best-performing models reported
in Table~\ref{tab:semisup_results} that achieved the best validation errors
after more careful hyperparameter tuning.

As in the proposed function $g$, all comparison denoising functions
mapped neuron-wise the corrupted hidden layer pre-activation $\tz l$ to
the reconstructed hidden layer activation given one projection from
the reconstruction of the layer above: $\hat z^{(l)}_i = g(\tilde
z^{(l)}_i, u^{(l)}_i)$.

\begin{table}[bth]
\begin{center}
 \begin{tabular}{lll}
   Test error \% with \# of used labels & 100 & 1000 \\
   \hline
   Proposed $g$: Gaussian $z$  & \textbf{1.06} ($\pm$ 0.07) & \textbf{1.03} ($\pm$ 0.06) \\
   \hline
   Comparison $g_1$: miniature MLP with $\tilde{z}u$ & 1.11 ($\pm$ 0.07) & 1.11 ($\pm$ 0.06) \\
   Comparison $g_2$: No augmented term $\tilde{z}u$   & 2.03 ($\pm$ 0.09) & 1.70 ($\pm$ 0.08) \\
   Comparison $g_3$: Linear $g$ but with $\tilde{z}u$ & 1.49 ($\pm$ 0.10) & 1.30 ($\pm$ 0.08) \\
   Comparison $g_4$: Only the mean depends on $u$     & 2.90 ($\pm$ 1.19) & 2.11 ($\pm$ 0.45)
 \\
\end{tabular}
 \end{center}
 \caption{Semi-supervised results from the MNIST dataset. The proposed function $g$~is compared to alternative parametrizations. Note that the hyperparameter search was not as exhaustive as in the final results,
 which means that the results of the proposed model deviate slightly from the final results presented
 in Table~\ref{tab:semisup_results}.}
 \label{tab:comparison_results}
\end{table}

The comparison functions $g_{1  \ldots 4}$ are parametrized as follows:

\subparagraph{Comparison $g_1$: Miniature MLP with $\tilde{z}u$}

\begin{equation}
	\hat z = g(\tilde{z}, u) = \mathbf{a} \boldsymbol{\xi} 
     + b \mathtt{sigmoid}(\mathbf{c}\boldsymbol{\xi})
	\label{eq:g_func_2}
\end{equation}
where $\mathbf{\xi}=[1, \tilde{z}, u, \tilde{z} u]^T$ is
an augmented input, $\mathbf{a}$ and $\mathbf{c}$ are trainable weight vectors, 
$b$ is a trainable scalar weight. This parametrization is capable
of learning denoising of several different distributions including sub-
and super-Gaussian and bimodal distributions. 

\subparagraph{Comparison $g_2$: No augmented term}

\begin{equation}
	g_2(\tilde{z}, u) = \mathbf{a} \boldsymbol{\xi}^\prime 
     + b \mathtt{sigmoid}(\mathbf{c}\boldsymbol{\xi}^\prime)
\end{equation}
where $\mathbf{\xi}^\prime=[1, \tilde{z}, u]^T$. 
$g_2$ therefore differs from $g_1$
in that the input lacks the augmented term $\tilde{z}u$.

% In the original
% formulation, the augmented term was expected to increase the freedom of the denoising
% to modulate the distribution of $z$
% by $u$. However, we wanted to test the effect on the results.

\subparagraph{Comparison $g_3$: Linear g}

\begin{equation}
	g_3(\tilde{z}, u) = \mathbf{a} \boldsymbol{\xi}.
\end{equation}
$g_3$ differs from $g$ in that it is linear and does not have a sigmoid term.
As this formulation is linear, it only supports Gaussian distributions.
Although the parametrization has the augmented term that lets $u$ modulate the slope
and shift of the distribution, the scope of possible denoising functions is still fairly limited.

\subparagraph{Comparison $g_4$: $u$ affects only the mean of $p(z \mid u)$}

\begin{equation}
g_4(\tilde{z}, u) = a_1 u + a_2 \mathtt{sigmoid}(a_3 u + a_4) +
a_5 \tilde{z} + a_6 \mathtt{sigmoid}(a_7 \tilde{z} + a_8) + a_9
\end{equation}
$g_4$ differs from $g_1$ in that the inputs from $u$ are not allowed to modulate the
terms that depend on $\tilde{z}$, but that the effect is additive. This means that 
the parametrization only supports optimal
denoising functions for a conditional distribution $p(z \mid u)$ where $u$ only
shifts the mean of the distribution of $z$ but otherwise leaves the
shape of the distribution intact.

\subparagraph{Results}

All models were tested in a similar setting as the semi-supervised fully connected MNIST
task using $N=1000$ labeled samples. We also reran the best comparison model on $N=100$ labels.
The results of the analyses are presented in Table~\ref{tab:comparison_results}.

As can be seen from the table, the alternative parametrizations of g
are inferior to the proposed parametrization, at least in the
model structure we use.
% A notable exception is the denoising function
% $g_5$ which corresponds to a Gaussian model of $z$. Although the
% difference is small, it actually achieved the best performance with
% both $N=100$ labels and $N=1000$ labels. It therefore looks like
% using two miniature MLPs, one for the mean and the other for the
% variance of a Gaussian denoising model, offers a slight benefit over a
% single miniature MLP that was used in the experiments in
% Section~\ref{sec:experiments}.

% Note that even if $p(z | u)$ is Gaussian given $u$, its marginal $p(z)$ is
% typically non-Gaussian. A conditionally Gaussian $p(z | u)$ which was
% implicitly used in $g_5$ forces the network to represent any non-Gaussian
% distributions with higher-level hidden neurons and these may well turn out
% to be useful features. For instance, if the marginal $p(z)$ is a mixture
% of Gaussian distributions, higher levels have a pressure to represent
% the mixture index because then $p(z | u)$ would be Gaussian and the denoising
% $g_5$ optimal as long as the higher layer represents the index information
% in such a way that $g_5$ can decode it from $u$.

These results support the finding by \citet{Rasmus15arxiv}
that modulation of the lateral connection from $\tilde z$ to $\hat z$ by
$u$ is critical for encouraging the development of invariant representations
at the higher layers of the model. Comparison function $g_4$ lacked this
modulation and it clearly performed worse than any other denoising function
listed in Table~\ref{tab:comparison_results}. Even the linear $g_3$ performed
very well as long it had the term $\tilde{z}u$. Leaving the nonlinearity but
removing $\tilde{z}u$ in $g_2$ hurt the performance much more.

In addition to the alternative parametrizations for the g-function, we ran experiments
using a more standard autoencoder structure. In that structure, we attached an additional
decoder to the standard MLP by using one hidden layer as the input to the decoder and
the reconstruction of the clean input as the target. The structure of the decoder was set
to be the same as the encoder: that is, the number and size of the layers from the input to the
hidden layer where the decoder was attached were the same as the number and size of
the layers in the decoder. The final activation function in the decoder was set to be the
sigmoid nonlinearity. During training, the target was the weighted sum of the reconstruction
cost and the classification cost.

We tested the autoencoder structure with 100 and 1000 labeled samples.
We ran experiments for all possible decoder lengths: that is, we tried attaching the decoder
to all the hidden layers. However, we did not manage to get a significantly
better performance than the standard supervised model without any decoder in any of the experiments.

\end{document}